\documentclass{IEEEoj}
\usepackage{cite}
\usepackage{amsmath,amssymb,amsfonts}
\usepackage{algorithmic}
\usepackage{graphicx,color}
\usepackage{textcomp}
\usepackage{verbatim}

\usepackage{siunitx}
\usepackage{url}
\usepackage{hyperref}
\usepackage{dirtytalk}
\usepackage{blindtext}
\usepackage{booktabs}
\usepackage{tabularx}

\usepackage[acronym]{glossaries}
\glsdisablehyper

\newacronym{mil}{MiL}{Model-in-the-Loop}
\newacronym{sil}{SiL}{Software-in-the-Loop}
\newacronym{hil}{HiL}{Hardware-in-the-Loop}

\def\BibTeX{{\rm B\kern-.05em{\sc i\kern-.025em b}\kern-.08em
    T\kern-.1667em\lower.7ex\hbox{E}\kern-.125emX}}
\AtBeginDocument{\definecolor{ojcolor}{cmyk}{0.93,0.59,0.15,0.02}}
\def\OJlogo{\vspace{-14pt}\includegraphics[height=28pt]{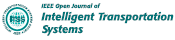}}

\begin{document}

\receiveddate{XX Month, XXXX}
\reviseddate{XX Month, XXXX}
\accepteddate{XX Month, XXXX}
\publisheddate{XX Month, XXXX}
\currentdate{XX Month, XXXX}
\doiinfo{OJITS.2022.1234567}

\title{EDGAR: An Autonomous Driving Research Platform -- From Feature Development to Real-World Application}

\author{
Phillip Karle$^{1}$,
Tobias Betz$^{1}$,
Marcin Bosk$^{2}$,
Felix Fent$^{1}$,
Nils Gehrke$^{1}$,
Maximilian Geisslinger$^{1}$,
Luis Gressenbuch$^{3}$,
Philipp Hafemann$^{1}$,
Sebastian Huber$^{1}$,
Maximilian Hübner$^{4}$,
Sebastian Huch$^{1}$,
Gemb Kaljavesi$^{1}$,
Tobias Kerbl$^{1}$,
Dominik Kulmer$^{1}$,
Sebastian Maierhofer$^{3}$,
Tobias Mascetta$^{3}$,
Florian Pfab$^{1}$,
Filip Rezabek$^{5}$,
Esteban Rivera$^{1}$,
Simon Sagmeister$^{1}$,
Leander Seidlitz$^{5}$,
Florian Sauerbeck$^{1}$,
Ilir Tahiraj$^{1}$,
Rainer Trauth$^{1}$,
Nico Uhlemann$^{1}$,
Gerald Würsching$^{3}$,
Baha Zarrouki$^{1}$,
Matthias Althoff$^{3}$,
Johannes Betz$^{6}$,
Klaus Bengler$^{4}$,
Georg Carle$^{5}$,
Frank Diermeyer$^{1}$,
Jörg Ott$^{2}$,
Markus Lienkamp$^{1}$
}%

\affil{School of Engineering and Design, Department of Mobility Systems Engineering, Institute of Automotive Technology and Munich Institute of Robotics and Machine Intelligence (MIRMI)}
\affil{School of Computation, Information and Technology, Department of Computer Engineering, Chair of Connected Mobility}
\affil{School of Computation, Information and Technology, Department of Computer Engineering, Professorship Cyber-Physical Systems}
\affil{School of Engineering and Design, Department of Mechanical Engineering, Chair of Ergonomics}
\affil{School of Computation, Information and Technology, Department of Computer Engineering, Chair of Network Architectures and Services}
\affil{School of Engineering and Design, Department of Mobility Systems Engineering, Professorship Autonomous Vehicle Systems and Munich Institute of Robotics and Machine Intelligence (MIRMI)}

\corresp{
All authors are with Technical University of Munich, Garching, GER

CORRESPONDING AUTHOR: Florian Pfab (e-mail: florian.pfab@tum.de)
}

\authornote{This work was supported by the Deutsche Forschungsgesellschaft (DFG)}

\markboth{Preparation of Papers for IEEE OPEN JOURNALS}{Author \textit{et al.}}


\begin{abstract}
While current research and development of autonomous driving primarily focuses on developing new features and algorithms, the transfer from isolated software components into an entire software stack has been covered sparsely.
Besides that, due to the complexity of autonomous software stacks and public road traffic, the optimal validation of entire stacks is an open research problem.
Our paper focuses on these two aspects. We present our autonomous research vehicle \textit{EDGAR} and its digital twin, a detailed virtual duplication of the vehicle.
While the vehicle's setup is closely related to the state of the art, its virtual duplication is a valuable contribution as it is crucial for a consistent validation process from simulation to real-world tests.
In addition, different development teams can work with the same model, making integration and testing of software stacks much easier, significantly accelerating the development process.
The real and virtual vehicles are embedded in a comprehensive development environment, which is also introduced. 
All parameters of the digital twin are provided open-source at \url{https://github.com/TUMFTM/edgar_digital_twin}.
\end{abstract}


\maketitle


\section{Introduction}

\IEEEPARstart{A}{utonomous}  software stacks must comprise a broad range of features to enter the complex environment of public road traffic.
The two major challenges from feature development to real-world application of these stacks are an efficient, early integration of isolated software components into an overall software stack and the optimal and complete validation of this full stack.
A holistic simulation environment is a crucial aspect to solve these challenges. This is due to the wide variety of scenarios~\cite{Li2019}, which require reproducible and easily scalable virtual testing to reduce the effort of real-world tests. Additionally, with a shared virtual development environment, integration effort is facilitated.
However, with available simulation platforms for autonomous vehicles (AV), there is a significant inconsistency between virtual and real-world tests because different models are used.
Thus, a digital twin, a virtual representation of the physical entity able to simulate the system~\cite{Souza2019}, is indispensable to ensure the consistency and validity of the results. The consistency of vehicle dynamics, sensor behavior, and network properties are the essential factors for this use case.
As Tao et al.~\cite{Tao2019} even state, creating a digital twin is one of the major challenges in the research and improvement of AVs.

\begin{figure}
    \centering
    \includegraphics[width=3.4in, clip, trim=0.0cm 2.4cm 4.5cm 2.3cm]{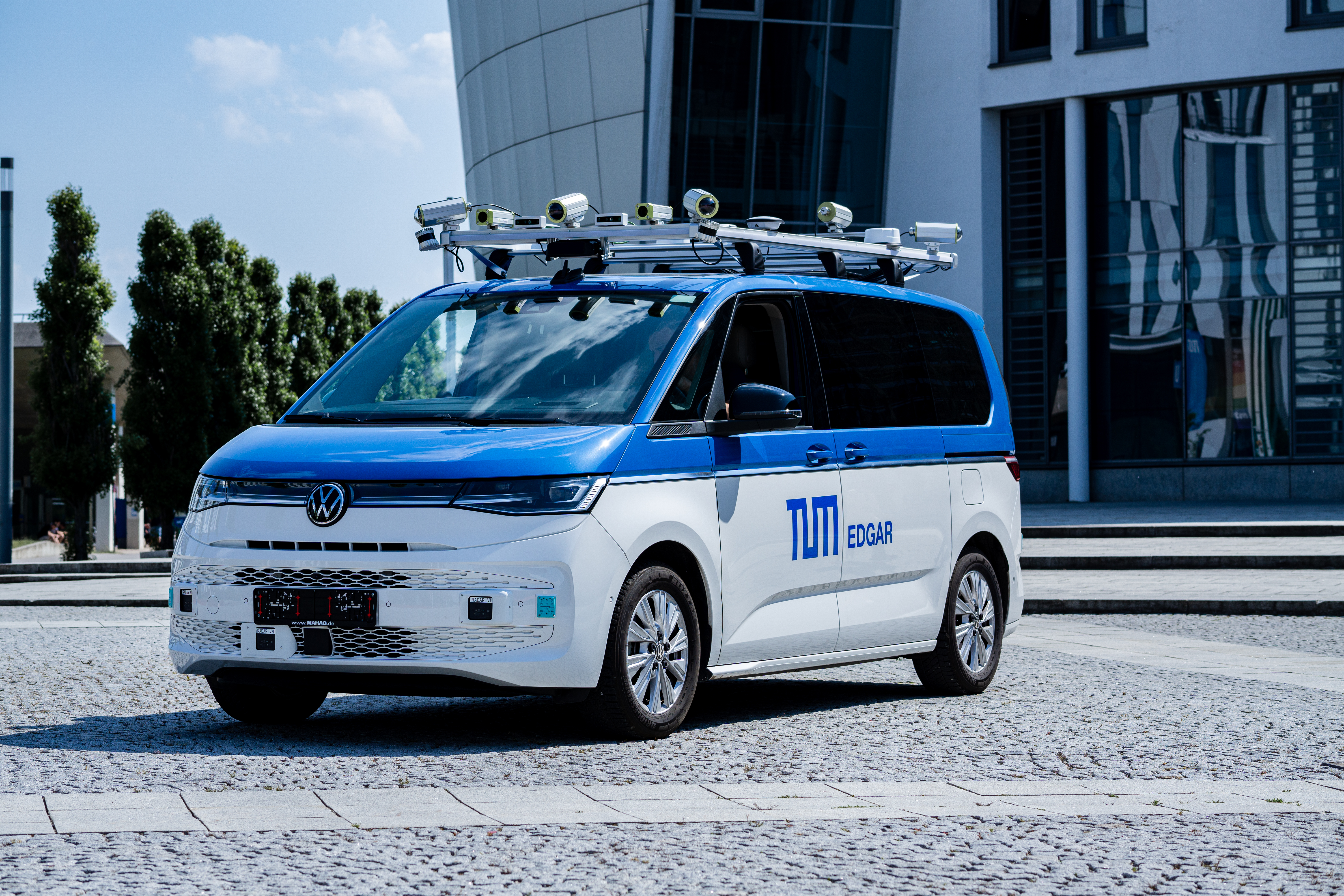}
    \caption{EDGAR: The research vehicle of the Technical University of Munich.}
    \label{fig:edgar_veh}
\end{figure}

The research platform presented in this article targets this aspect: We propose our research vehicle called \textit{EDGAR} (Excellent Driving GARching, Fig. \ref{fig:edgar_veh}) and its digital twin, a detailed virtual duplication.
In summary, the main contributions of this paper are:

\begin{itemize}
    \item We propose an autonomous research vehicle with a multi-sensor setup and different computing hardware architectures (x86, ARM) that addresses multiple research topics (perception, planning, control, teleoperation, HMI, network communication, V2X).
    \item We present and validate a comprehensive digital twin with vehicle dynamics models and sensor and network replication for a consistent testing strategy from simulation to reality. The digital twin setup is available open-source. To the best of our knowledge, this is the first publicly available digital twin of an autonomous road vehicle.
    \item We introduce a holistic workflow starting from feature development over multiple simulation steps to real-world testing in which the real and virtual vehicles are embedded. In addition, the development environment offers a large-scale data center for systematic data handling of stored sensor data and software logs to foster software development.
\end{itemize}


\begin{figure*}
\centerline{\includegraphics[width=1.95\columnwidth]{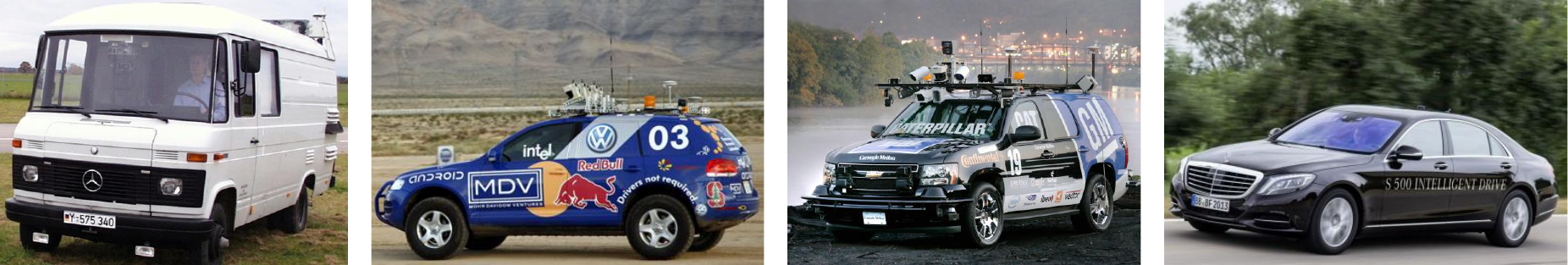}}
\caption{Milestones of AV research vehicles: VaMoRs  \cite{Maurer1995}, Stanley  \cite{Thrun2006}, Boss  \cite{Urmson2008}, MB S-Class  \cite{Ziegler2014}}
\label{fig:av_cars}
\end{figure*}

\section{Related Work}
\label{sec:related_work}
The following section covers the related work of AV development systems. These systems comprise research vehicles for real-world testing and data recording and evaluation processes. For a general survey about the state of the art of hardware and software for AVs, we refer to  \cite{Yurtsever2020, Badue2021}.

Since the 1980s, prototypical vehicles have been set up to demonstrate the capabilities of autonomous systems in real-world traffic  \cite{Dickmanns2002}. Among these are Alvinn  \cite{Pomerleau1988}, a camera-end-to-end single-lane following vehicle, VaMoRs  \cite{Maurer1995}, a computer vision-based vehicle for lateral and longitudinal guidance, Prometheus  \cite{Williams1988} with a 4D-based approach of image-processing. Fig.~\ref{fig:av_cars} shows some of these iconic cars. Based on these milestones, Table~\ref{tab:res_vehicles} outlines research vehicle platforms. It can be observed that most work on AV research vehicles focuses on the real vehicle setup, including sensors and computing platforms. None of them offers a digital twin of the presented vehicle.

\begin{table*}
  \footnotesize\sf\centering
  \setlength{\tabcolsep}{0.19cm}
  \begin{center}
    \caption{Overview of the state of the art of AV research vehicles.}
    \label{tab:res_vehicles}
    \centering
    \begin{tabular}{lccccccccc}
    \toprule
    & \multicolumn{3}{c}{\textbf{Environment Sensors}} & \multicolumn{3}{c}{\textbf{Computing Platform}}  & \multicolumn{3}{c}{\textbf{Digital Twin}}   \\
    & Camera & LiDAR & RADAR & PC & Real-Time & GPU & Sensors &  Vehicle Dynamics  & Network \\
      \midrule
      Stanley \cite{Thrun2006} & 1 & 5 & 2 & \checkmark & & & \\ 
      Junior \cite{junior2008, Levinson2011} & 0 & 8 & 5 & \checkmark & & & \\  
      Boss \cite{Urmson2008} &  2 & 11 & 5 & \checkmark & & &  \\ 
      Bertha \cite{Ziegler2014} & 3 & 0 & 8  & & & & & \\ 
      Braive \cite{Grisleri2010} & 10 & 5 & 0 & \checkmark & & & \\ 
      CMU vehicle \cite{Wei2013} & 2 & 6 & 6 & \checkmark & & \checkmark  & & \\  
      Fortuna \cite{Buechel2019, Kessler2019} & 5 & 3 & 4 & \checkmark & \checkmark & \checkmark & & \\ 
      Jupiter \cite{Haselberger2022} & 1 & 3 &  & \checkmark & \checkmark  & & \\  
      X-Car  \cite{Mehr2022} & 7 & 1 & 1 & \checkmark & & \checkmark  & & \\  
      \hline
      \textit{EDGAR} (ours) & 10 & 4 & 6 & x86, ARM & ARM & \checkmark & \checkmark & \checkmark & \checkmark 
      \\ 
      \bottomrule
    \end{tabular}
  \end{center}
\end{table*}

Several works focus on the requirements for an AV sensor setup~\cite{Broggi2010, Chen2012} and how to fulfill them~\cite{Cho2014, Brummelen2018, Leonard2008, Taraba2018, Campbell2018, Furgale2013}.
Furthermore, various research institutions use vehicles with sensors and computation hardware to gather data from road scenarios~\cite{Geyer2020, Ligocki2020, Burnett2022, Carballo2020}. However, these vehicles are not used for testing and evaluating self-driving software.

A reliable development workflow is required to test and validate the developed AV features. A common approach is to start on the feature and module level with unit tests and Model-in-the-Loop (MiL), followed by overall software tests in Software-in-the-Loop (SiL) and Hardware-in-the-Loop (HiL) simulations. Afterward, the real-world tests are the final stage to validate the software.
Gao et al.~\cite{Gao2022} propose such an evaluation system and provide an overview of the required infrastructure and methods for this system. They name four major research directions to focus on: Virtual reality-based driving scenarios, automated safety scenario validation, reliable machine learning analysis methods, and system security evaluation models.
Thorn et al.~\cite{thorn2018framework} present a scenario-based test framework for automated driving functions of SAE Level~3-5~\cite{SAELEVEL}. In addition, fail-operational and fail-safe strategies are identified for the investigated AV systems.
Similarly, Chakra~\cite{Chakra2022} analyses the sim2real-gap for resilient real-world applications. The author states the three main areas for future research: The definition and measurement of AV intelligence, the general enhancement in AV simulation frameworks and methodologies, and AV simulation transferability and integration.
With a focus on real-time capability and hardware constraints for mobile applications, Lin et al.~ \cite{Lin2018} present and formalize a design guideline. However, there is no analysis of the AV software features.
A practical framework for an architecture design of software and hardware to create an overall AV system is shown in~\cite{Zong2018}. Their contributions are comparing different sensor setups, a complete AV software validated in real world, and new scalable data transmission systems.

The evaluation and testing process can also be bi-directional, i.e., new development requirements and data for feature development can be derived during the evaluation and testing process.
The extraction of information from real-world tests for further feature development is demonstrated by Liu et al.  \cite{Liu2020}. The proposed pipeline uses logs from real-world tests for software feature development. They focus their development pipeline on self-adaptive path planning. The proposed planning algorithm is able to improve its knowledge base of collision scenarios after a test is performed and thereby avoids dangerous situations that might occur during testing in the future. In contrast, Deliparaschos et al.  \cite{Deliparaschos2020} propose to derive data from sensors placed on the road infrastructure to extract driving scenarios for model verification and validation.
Zhao et al.  \cite{Zhao2017} extend the idea of real-world data collection by data augmentation. They propose to collect data from real-world driving, focusing on real-world edge case scenarios, i.e., meaningful and safety-critical scenarios. A Monte Carlo simulation is applied to these scenarios for higher complexity. Subsequently, the simulation enables to derive a statistical analysis of how the AV would perform in everyday driving conditions. The authors conclude that by means of this method, the real-world effort can be significantly reduced, but the variety of scenarios remains high due to the synthetic augmentation.

Regarding the quantification of the test effort, Hauer et al.~\cite{Hauer2019} propose a method to determine if an AV system is tested in all scenarios, i.e., if sufficient real drive data is collected. Using the Coupon Collector's problem, a statistical guarantee can be given that all scenario types are covered.
The standardization of AV performance evaluation is the focus of the work of Basantis et al.  \cite{Basantis2019}. They conclude that standardized testing is a valuable tool to evaluate the capabilities of AV vehicles and that a robust evaluation mechanism greatly impacts AV systems' conformance.

In summary, the presented work on AV research vehicles either focuses on the hardware setup or states the deployment method for the software without validation. 
The work in the field of AV evaluation and development systems is mainly conceptual, i.e., only requirements are identified for AV software validation.
Even though the process to base feature development on test data is introduced in the state of the art, the implementation of an AV validation concept in an overall software development workflow is sparsely covered, e.g., by  \cite{thorn2018framework}. 
The consideration of a digital twin, which connects the real-world system with the virtual one and thus ensures the consistency and validity of the entire process, is completely neglected.

Our aim is to establish this consistency with our research vehicle \textit{EDGAR} and its digital twin, embedded in our proposed workflow from feature development to real-world application.
In the scope of this work, a detailed introduction of the hardware setup of our AV research vehicle is given.
The related digital twin comprises vehicle dynamic models and sensor and network replication, which are also presented.
Our development workflow comprises the forward step from feature development via multi-stage testing and validation to real-world application and the backward step to use the real-world data for the improvement of the simulation and to derive new feature requests. 


\section{Autonomous Vehicle Setup}
\label{sec:av_setup}
The hardware setup of \textit{EDGAR} is described in the following section. Starting with the base vehicle (\ref{subsec:veh}), we then introduce the sensors mounted on the vehicle (\ref{subsec:sensors}) and the computer and network components (\ref{sec:com_netw}). For each component, the design decisions from a technical point of view and the constraints for the final decision are described. The description of the actuation interfaces completes the overview of the vehicle hardware setup and the actuation interfaces (\ref{sec:actuation}). Finally, the HiL simulation framework (\ref{sec:hil_sim}) and the data center (\ref{sec:datacenter}) are described. Table~\ref{tab:hw_overview} lists all hardware components.

\begin{table}
\caption{Overview of the hardware components (MR: Mid-Range, LR: Long-Range, GM: Grandmaster).}
\label{tab:hw_overview}
\setlength{\tabcolsep}{3.0pt}

\centering
\begin{tabular}{llll}
\toprule

\textbf{Component}& 
\textbf{Producer}&
\textbf{Model}
\\
\midrule

Series vehicle&
Volkswagen& 
T7 Multivan 1.4 eHybrid\\

Mono camera&
Basler& 
acA1920-50gc\\

Stereo camera&
Framos&
D455e\\

LiDAR MR&
Ouster& 
Ouster OS1-128\\

LiDAR LR&
Innovusion& 
Falcon Gen-2\\

RADAR&
Continental& 
Continental ARS430\\

GPS-IMU&
Novatel& 
PwrPak7D-E2\\

Microphones&
Infineon&
A2B Eval Kit\\

x86 HPC&
InoNet& 
Mayflower B17
\\

ARM HPC&
ADLINK& 
AVA AP1\\

Network Switch&
Netgear&
M4250-40G8XF-PoE+\\

PTP GM&
Masterclock& 
GMR5000\\

SDR Transceiver&
Ettus&
USRP B210 SDR Kit 2x2\\

V2X System&
Cohda&
MK5 OBU\\

5G-Router&
Milesight&
UR75-500GL-G-P-W\\

MIMO Antennas&
Panorama&
LGMQM4-6-60-24-58 4x4\\

LED&
BTF lighting&
WS2815\\

LED controller&
Strip Studio&
SPI Matrix\\

Visualization PC&
Spo-comm&
RUGGED GTX 1050 Ti\\


\bottomrule
\end{tabular}
\end{table}

\subsection{Vehicle} \label{subsec:veh}
The Volkswagen T7 Multivan Style 1.4 eHybrid is a hybrid electric vehicle and the basis for the research vehicle \textit{EDGAR}.
One of the key advantages of using the T7 Multivan Style 1.4 eHybrid is its hybrid powertrain. The vehicle has a 1.4-liter turbocharged four-cylinder engine, an electric motor, and a \SI{13}{\kWh} lithium-ion battery.
With an estimate of about \SI{1.5}{\kilo\watt} continuous power, the vehicle's alternator provides enough power to operate prototyping computers and sensors. Another benefit of the Volkswagen T7 is its advanced features, such as Adaptive Cruise Control, Lane Departure Warning, Park Assist, and Emergency Assist. These features provide safety-certified actuator interfaces that can be reused cost-effectively as a fallback for the developed software and comparison. Additionally, the space inside is advantageous to placing all components easily accessible in the trunk and improving airflow to cool them.
A computer is placed between the two front seats with two screens mounted inside the vehicle to visualize the state of the AV software and vehicle during test rides.

Despite the advantages of using the T7 Multivan Style 1.4 eHybrid as a research platform, there are also some limitations to consider. The vehicle is relatively large, which can make it less suitable for testing in densely populated urban areas. The vehicle's height especially poses difficulties in sensor placement, with a trade-off between near-field coverage and a far-range sensor focus.

\subsection{Sensors}
\label{subsec:sensors}
Our sensor setup consists of cameras, LiDARs, RADARs, and microphones for perception of the local environment. In addition, a GPS-IMU system is included. Two requirements that apply to the whole sensor setup are:
\begin{itemize}
    \item Precision Time Protocol (PTP) capability for time synchronization inside the car
    \item ROS2-compatible drivers for software integration
\end{itemize}
Further requirements, intended use cases, and final choice for each sensor are discussed in the following subsections.

\begin{figure}[!t]
    \centering{\includegraphics[clip, trim=0.0cm 0.0cm 0.0cm 6.0cm, width=0.99\columnwidth]{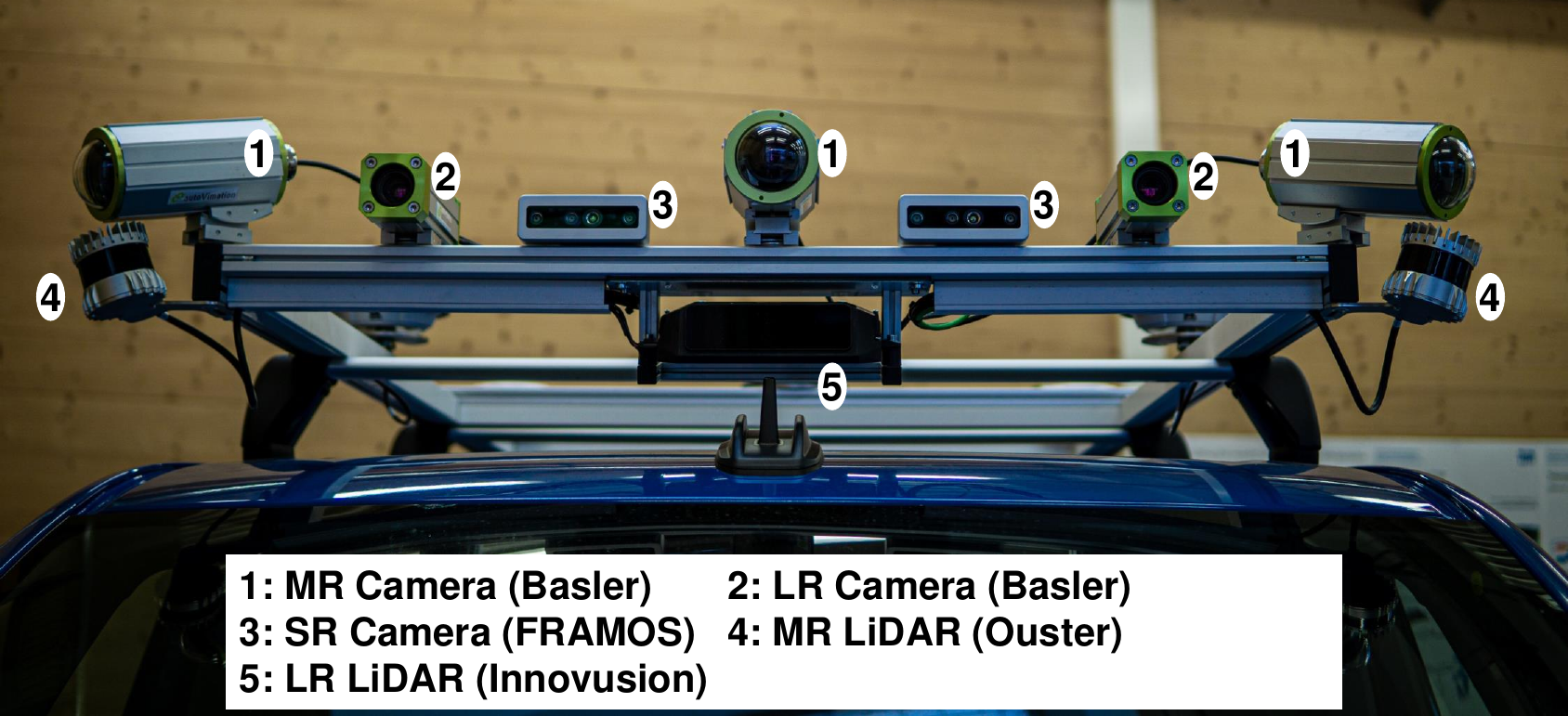}}
    \caption{Front roof area with LiDAR and Camera sensors (MR: Mid-Range, LR: Long-Range, SR: Short-Range).}
    \label{fig:edgar_front_roof}
\end{figure}

\begin{figure*}
\centerline{\includegraphics[width=2.0\columnwidth]{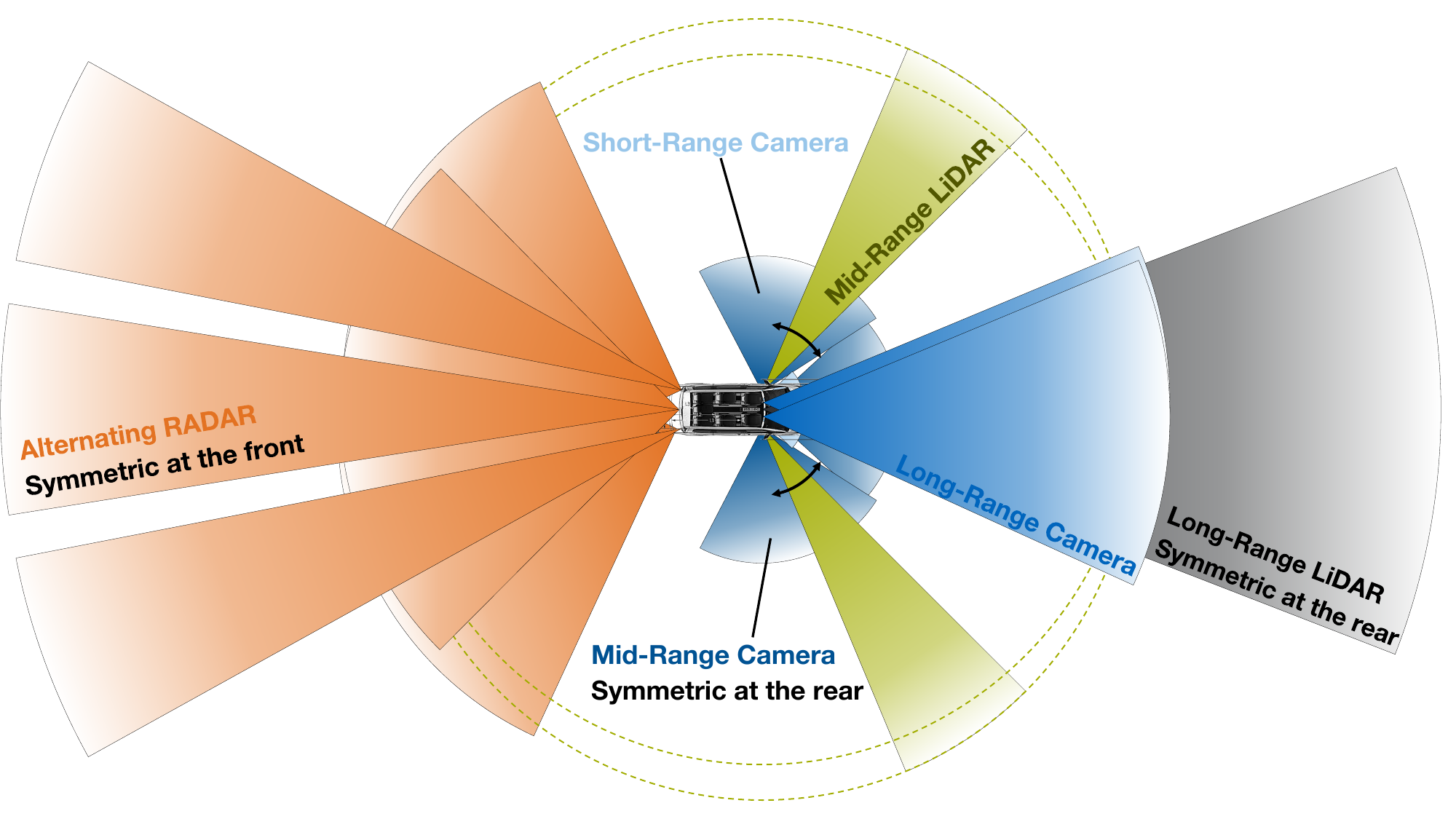}}
\caption{
Field of view of the perception sensors.
}
\label{fig:bev_sensors_perception}
\end{figure*}

The described setup enables holistic coverage with camera, RADAR, and LiDAR sensors. Fig. \ref{fig:edgar_front_roof} depicts a front view of the sensor roof rack showing the front LiDARs and cameras, and Fig. \ref{fig:bev_sensors_perception} shows the field of view of the perception sensors.
The detailed positions and orientations of the sensors are given in the repository of the digital twin. All positions and orientations are stored in a .urdf-file. The reference point is the middle of the rear axle.

\subsubsection{Camera}
The camera setup of the research vehicle was designed with two applications in mind: Autonomous driving and vehicle teleoperation. For these two applications, the following requirements were identified:
\begin{itemize}
    \item \SI{360}{\degree} Field-of-View (FoV); 
    \item a minimal resolution of 1280 x 720 pixel;
    \item a frame rate of \SI{40}{fps} to minimize the latency;
    \item a consistent camera-lens combination to simplify stitching the camera images; and
    \item the option for depth completion through stereo cameras.
\end{itemize}
\textbf{Mid-Range:}
To fulfill these criteria, we chose 6 Basler acA1920-50gc cameras with the Sony IMX174 CMOS sensor for a \SI{360}{\degree} representation with mono camera images, 
These cameras are capable of providing Full HD (1920 x 1200 pixel) color images at a maximum frame rate of \SI{50}{\hertz}.
Three cameras are mounted on the front center and corners of the vehicle's roof.
Combined with lenses with a focal length of \SI{6}{\milli\meter} (Kowa LM6HC), each front camera provides a horizontal FoV of \SI{84.9}{\degree} and a vertical FoV of \SI{59.7}{\degree}.
At the rear end of the vehicle's roof, three cameras are mounted in combination with a Kowa LM4HC lens, having a focal length of \SI{4.7}{\milli\meter}.
The resulting FoVs (horizontal: \SI{99.5}{\degree}, vertical \SI{73.1}{\degree}) minimize the blind spots at the vehicle's sides.
\\
\textbf{Long-Range:}
Two additional Basler acA1920-50gc cameras are mounted at the front of the vehicle. These cameras are combined with lenses having a focal length of \SI{16}{\milli\meter}, resulting in a horizontal FoV of \SI{38.6}{\degree} and a vertical FoV of \SI{24.8}{\degree}. The purpose of these cameras is to provide stereovision and far-range vision.
\\
\textbf{Short-Range:}
Two FRAMOS D455E depth cameras are attached to the vehicle's roof rack. These cameras make use of the active IR Stereo technology and provide depth images with a maximal resolution of 1280 x 720 pixels at a maximal framerate of \SI{30}{\hertz}.

\subsubsection{LiDAR}
\begin{figure}
\centerline{\includegraphics[width=0.99\columnwidth]{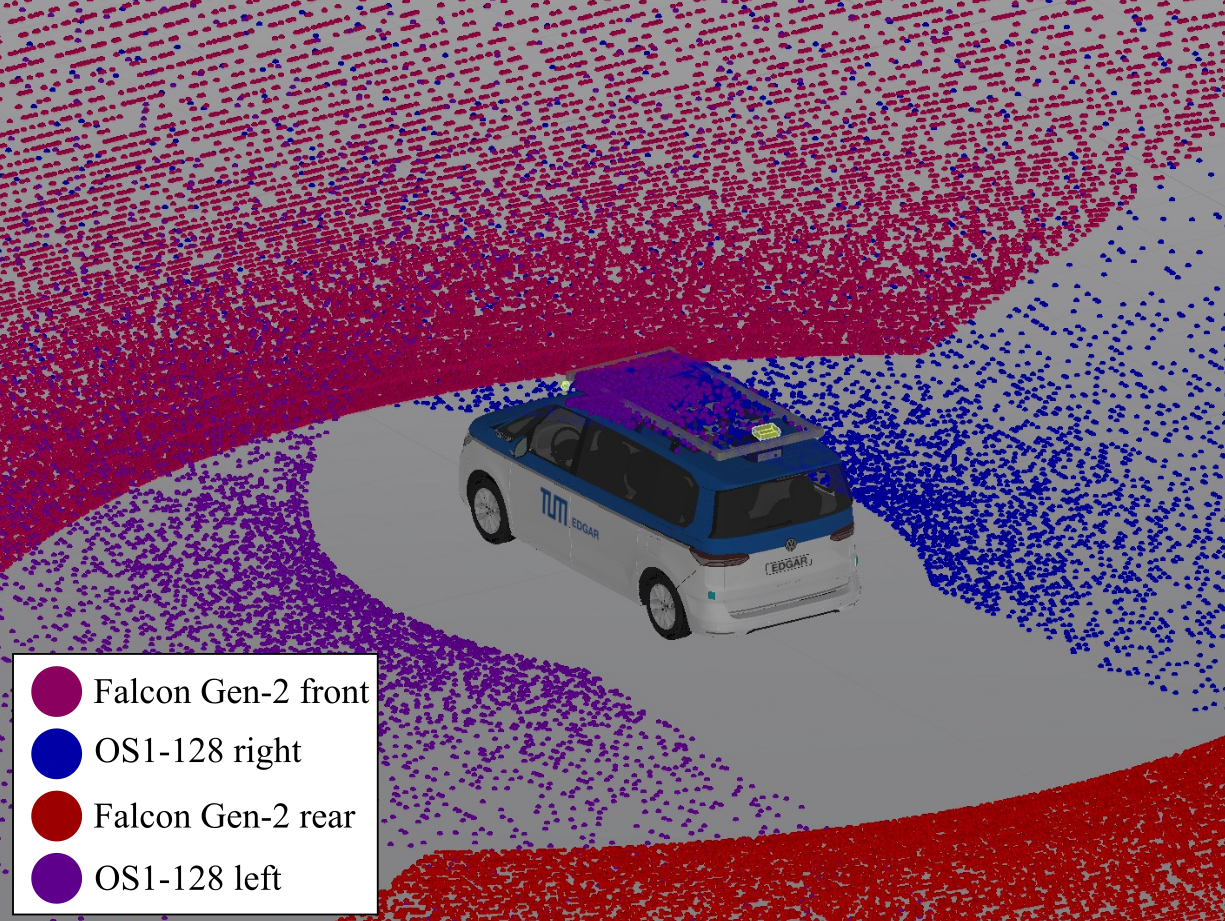}}
\caption{Unity environment to simulate point clouds for different LiDAR configurations. \label{fig:lidar_unity}}
\end{figure}

The LiDAR setup was designed to enable autonomous driving in arbitrary traffic scenarios (that is, urban and highway settings). In summary, our requirements are the following:
\begin{itemize}
    \item \SI{360}{\degree} FoV with blind spots as small as possible directly around the vehicle;
    \item high range, especially to the front and rear; and
    \item dense reflections in close and mid-range.
\end{itemize}
For the selection process of a LiDAR setup, we created a simulation environment based on \textit{Unity} to generate synthetic point clouds with different setups.
The simulation was based on \cite{Betz_2023} and adapted to the new base vehicle.
An exemplary simulation of a LiDAR setup is shown in Fig.~\ref{fig:lidar_unity}. It can be seen that a setup of four LiDAR sensors satisfies our needs of minimized blind spots around both sides of the car but a high perception range to the front and rear best.
Two rotating sensors are placed on the vehicle roof's left and right front corners.
Two long-range solid-state LiDARs are placed at the front and rear center of the roof.

\textbf{Mid-Range}:
For short- and mid-range detections, two Ouster OS1-128 were chosen.
These are rotating \SI{360}{\degree} sensors with a vertical FoV of \SI{45}{\degree} with higher resolution in the center of the vertical FoV.
They offer a range of \SI{45}{\meter} at \SI{10}{\percent} reflectivity using a \SI{865}{\nano\meter} wavelength laser.
\\
\textbf{Long-Range}:
To cover the most important regions around the vehicle, namely front and rear, two additional LiDARs were added to the corresponding centers of the roof.
These put more emphasis on long-range detections.
Therefore, the chosen Innovusion Falcon offers a range of \SI{250}{\meter} at \SI{10}{\percent} reflectivity through \SI{1550}{\nano\meter} laser.
They have an FoV of \SI{120}{\degree} horizontally and \SI{25}{\degree} vertically.
Since the scan pattern of these LiDARs is software-defined, regions of interest (ROI) can be defined at runtime to locally increase the resolution.

\subsubsection{RADAR}
In addition to LiDAR and camera, we also use RADAR sensors because of their high robustness against severe weather conditions and their ability to measure the velocity of the target objects. The sensors are used for RADAR-only detection and fusion algorithms, e.g., Camera-RADAR-Fusion. The requirements to choose an appropriate sensor were the following:
\begin{itemize}
    \item High detection range for long-range perception compared with a broad horizontal field of view for short-range perception;
    \item accurate measure of the velocity of target objects;
    \item high point cloud density; and
    \item 3D-detection, incl. elevation measurement.
\end{itemize}
Based on these requirements, our research vehicle's final choice is the Continental ARS430 RADAR sensor. Six of them are mounted on the vehicle in total. These pulse compression RADAR modulation sensors operate in the \SI{77}{\giga\hertz} frequency band and alternate between a far- and a near-field scanning pattern, with a horizontal field of view of \SI{\pm9}{\degree} and \SI{\pm60}{\degree}, respectively. The sensors have a maximum detection range of \SI{250}{\meter} and an azimuth angular resolution of \SI{0.1}{\degree}. RADAR sensors can measure the velocity of target objects with an accuracy of \SI{0.1}{\km\per\hour} and detect objects up to a minimum RADAR Cross Section of \SI{10}{\metre\squared} (in the range of \SI{200}{\meter}). In our case, the sensor has primarily been selected due to its ability to output data in a point cloud data format and its information density (high number of output points). Finally, the possibility of PTP time synchronization and integration into a ROS2 framework were two other important decision criteria. In contrast, the sensor misses the ability to measure elevation information and is limited to a \SI{100}{\mega\bit\per\second} BroadR-Reach Ethernet connection, which restricts the number of output points.

\subsubsection{GPS-IMU}
\label{sec:gpsimu}
A Global Navigation Satellite System (GNSS) system is used to locate the research vehicle globally. The following requirements were identified:
\begin{itemize}
    \item Combined system of GNSS and Inertial Navigation System (INS);
    \item support of Real-Time Kinematic (RTK); and
    \item measurement of vehicle heading at standstill.
\end{itemize}
Based on these requirements, we decided to use the NovAtel PwrPak7D-E2, a combined GPS-IMU system. The device supports Real-Time Kinematic (RTK) \cite{Hatch1991}, which allows receiving GNSS correction data over the Internet. As a result, it allows us to determine the vehicle's current location with an accuracy of $\SI{2}{\centi\meter}$ on average\footnote{\raggedright\url{https://hexagondownloads.blob.core.windows.net/public/Novatel/assets/Documents/Papers/PwrPak7D-E2-Product-Sheet/PwrPak7D-E2-Product-Sheet.pdf} \label{footn:hexagon}}. Furthermore, it can measure the vehicle's current heading at standstill by using two NovAtel GNSS-850 antennas. This is essential if the autonomous vehicle intends to start driving autonomously after being booted. The integrated Inertial Measurement Unit (IMU) is used to derive the vehicle's dynamic state during a GNSS outage until the vehicle can do a safe stop. One disadvantage of the PwrPak7D-E2 is that it does not yet support PTP. As a result, the Pulse-per-Second (PPS) output of the system must be used for time synchronization\footref{footn:hexagon}.

\subsubsection{Microphones}

While most autonomous research vehicles only use cameras, RADAR, and LiDAR sensors for their perception pipeline, we decided to also use microphones.
This enables further use cases like detection and localization of emergency vehicle sirens, blind spot detection, and road surface type estimation~\cite{Siwek2021}.
Our most important requirements are:
\begin{itemize}
    \item High fidelity;
    \item resistance to adverse environmental conditions;
    \item small physical dimensions; and
    \item low power consumption.
\end{itemize}

The XENSIV\texttrademark{} IM67D130A MEMS microphones from Infineon Technologies offer a signal-to-noise ratio (SNR)~$\geq~\SI{67}{\decibel}$ for improved audio quality and an acoustic-overload-point (AOP) $\geq \SI{130}{\decibel}$ for high wind-noise robustness.
Their housing is IP68-certified to protect the microphones from rain and dust.
Infineon's A\textsuperscript{2}B evaluation kit offers an AURIX microcontroller as an ECU master unit and four slave modules with four microphones each.
The slave modules are attached to the vehicle corners and connected via A\textsuperscript{2}B audio bus from analog devices.

The AURIX microcontroller can be flashed with custom code for audio preprocessing.
It is connected to the High Performance Computing (HPC) platform via Ethernet, where detection and localization tasks can be executed.

\subsection{Computer and Network}
\label{sec:com_netw}
The computer and network system comprises two high-performance computers, a network switch and a PTP grandmaster (GM), introduced in the following subsections.

\subsubsection{High-Performance Computer} \label{sec:hpc} The research vehicle is equipped with HPC platforms to handle the processing demands of autonomous driving software. The HPC platforms should satisfy the following requirements:

\begin{itemize}
    \item Multi-core (16) CPU with high clock frequency and large RAM for an overall low software latency \cite{Betz2022IAC};
    \item a high GPU-capacity to run deep learning applications;
    \item a CAN interface to the vehicle actuators;
    \item a high network bandwidth to receive the data; and
    \item a suitable storage setup for data recording.
\end{itemize}

\begin{table}[b!]
\caption{Specification of the vehicle computing platforms.}
\label{tab:hpc_description}
\centering
\setlength{\tabcolsep}{0.5em}
\begin{tabular}{lll}
\toprule
& \textbf{InoNet Mayflower-B17} & \textbf{ADLINK AVA AP1} \\
\midrule
\textbf{Architecture} & x86 & AArch64 \\
\textbf{CPU} & AMD EPYC 7313P & Ampere Altra Q80-26 \\
\textbf{Cores/Threads} & 16/32 & 80/80 \\
\textbf{Clock Frequency} & \SI{3.0}{\giga\hertz} (max \SI{3.7}{\giga\hertz}) & \SI{2.6}{\giga\hertz} \\
\textbf{RAM} & \num{4} x \SI{32}{\giga\byte} DDR4& \SI{96}{\giga\byte} DDR4\\
\textbf{Safety Island} & -- & NXP S32S247TV \\
\textbf{Network Interface} &  \multicolumn{2}{c}{SFP+ \num{4} x \SI{10}{\giga\bit\per\second}} \\
\textbf{GPU}  & \multicolumn{2}{c}{NVIDIA RTX A6000  \SI{48}{\giga\byte}} \\
\textbf{Operating System}  & \multicolumn{2}{c}{Ubuntu 22.04 Jammy Jellyfish} \\ 
\bottomrule
\end{tabular}
\end{table}

In addition, we use two different HPC platforms based on the x86 and aarch64 architecture to compare their performance for autonomous driving systems. Based on the requirements, we use two platforms: the x86-based InoNet Mayflower-B17 and the ARM-based ADLINK AVA AP1. The specifications are given in Table~\ref{tab:hpc_description}. A first evaluation of the x86-HPC processing power for Autoware.Universe can be found in \cite{betz2023latency}. We deactivate the hyper-threading option at the x86-HPC, which leads to lower latencies in the software stack. Besides the selected GPUs, we integrate the FPGA-based development board AMD VCK5000, which enables fast AI inference. 
Both platforms have built-in connectivity options for communication, such as CAN interfaces to access the vehicle actuators. 
Since a large amount of data is recorded with the research vehicle, we use Mayflower quick trays. These trays enable a fast change of the \num{4} x \SI{2}{\tera\byte} NVMe SSDs.
The AVA AP1 platform features the integrated safety island, which is an additional high-safety, real-time capable CPU to execute safety-critical functions. 
The aforementioned architecture allows the pursuit of new research topics, such as the development of safety features to ensure that the autonomous system can continue to operate even in the event of component failure.

\subsubsection{Network switch}
The network switch is a central component of the AV hardware setup. For the given use case, the following requirements were identified:
\begin{itemize}
    \item Data transfer from all sensors (downlink)
    \item Data transfer to the HPCs (uplink) via Power-over-Ethernet (PoE)
    \item Audio-Video-Bridging (AVB) and IEEE 802.1Qav
    \item PTP compatibility
\end{itemize}
The chosen device, M4250-40G8XF-PoE+ by Netgear, fulfills these requirements. It offers 40 PoE+ ports and 8 SFP+ ports, which are patched to \num{2} x \SI{40}{\giga\bit\per\second} uplink, one for each HPC, and supports Audio Video Bridging (AVB), IEEE 802.1Qav, and additional Time-Sensitive Networking (TSN) standards. In addition, the network switch serves as a transparent clock in the cascaded PTP system, i.e., it modifies the PTP timestamp from the PTP GM based on its residence time.

\subsubsection{PTP Clock Synchronization}
Time synchronization is essential in a system with multiple different and distributed sensors to record high-quality data without a time shift between individual sensors. Such capabilities are even more prominent when multiple vehicles exchange information and require freshness of the data. The requirements for this component are the following:
\begin{itemize}
    \item PTP master functionality to synchronize the system clocks of several PTP slaves in the network~\cite{IEEE1588};
    \item Synchronization with non-PTP-capable devices; and
    \item low clock drift.
\end{itemize}
PTP is organized in a master-slave hierarchy, where the slave device always synchronizes its internal clock with the information provided by the master.
For that, the PTP standard defines three clock types - boundary clock (BC), ordinary clock (OC), and transparent clock (TC). 
The OC has only a single port that is either in a master or slave state. 
On the other hand, the BC has two or more ports and is used to link complex PTP topologies. 
Last is the TC, which is not a master or slave and does not have an internal clock.
TC forwards PTP messages, adjusts their time correction field according to the residence time, and improves the PTP synchronization precision \cite{RezHelm22}. 
On top of this hierarchy is the GM clock that determines the clock for the whole system.

For our setup, we chose the Masterclock GMR5000, a state of the art PTP GM clock.
This allows maximum modularity and flexibility in switching components compared to a solution where, for example, the GNSS System would act as GM.
Furthermore, the GMR5000 offers multiple interfaces for time-synchronization with non-PTP-capable devices (\ref{sec:gpsimu}), increasing the number of potentially usable system components and sensors.

The clock can be synchronized with the current GNSS time by connecting the GMR5000 to a GNSS antenna or receiving the PPS signal from the vehicle's GNSS system.
The GMR5000 is equipped with an optional high-stability oscillator that lowers the time drift to about $\pm\SI{0,25}{\second\per year}$ \footnote{\raggedright\url{https://static1.squarespace.com/static/55f05c0ce4b03bbf99b13c15/t/5e8b6093a17e09405bb5e7ea/1586192532769/GMR5000+Data+Sheet.pdf}}. Therefore, time offsets between GNSS time and the vehicle's time stay low which simplifies time synchronization during the boot phase of the car, especially after prolonged phases of vehicle shutdown.
For the \textit{EDGAR} vehicle, we do not need a clock synchronized to, e.g., GPS, but it is crucial to have such capabilities when the vehicle communicates to external parties. 

To distribute the GM messages, we synchronize the Ego Vehicle (x86 HPC) with the GM.
The x86 HPC serves as BC as it has multiple ports and allows for good interconnectivity. 
Besides, it is connected to the Netgear switch that runs in the TC mode and introduces less clock jitter. 
Therefore, there are only two hops from the x86 to the sensors. 
The sensors operate in the OC mode and listen to the clock information provided by the x86 HPC. 
We chose the x86 HPC as a master device for the sensors as it provides more granular control of the PTP configuration without sacrificing clock precision. 

\begin{figure*}[ht]
\centerline{\includegraphics[width=\textwidth]{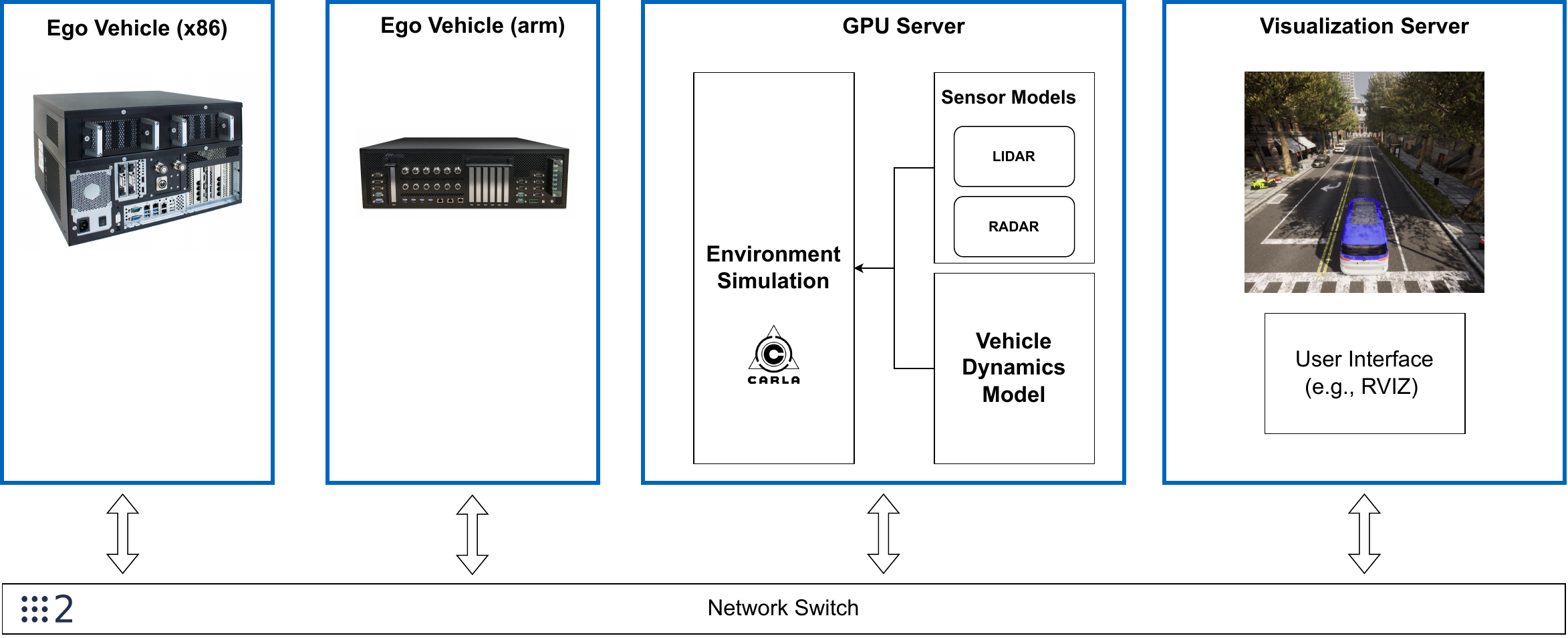}}
\caption{Overview about the HiL simulation setup. \label{fig:hil_setup}}
\end{figure*}

\subsubsection{External Communication}
The design choices for our external communication system aim to increase usage flexibility for different communication technologies. This includes software-defined radio (SDR) transceivers, a vehicle-to-everything (V2X) system, a router, and multiple-input multiple-output (MIMO) antennas. 

SDR is a communication system where several parts of the communication functionality can be configured in software \cite{Macedo2015}. 
We integrated three Ettus USRP B210 SDR Kit transceivers.
For communication with roadside infrastructure and other vehicles, we use the Cohda Wireless MK5 OBU\footnote{\url{https://www.cohdawireless.com/solutions/hardware/mk5-obu/}} V2X system, which supports the IEEE 802.11p V2X communication standard.

The vehicle's internet connection, which is received via 5G standard and allows communication to infrastructure, cloud-based computing, and teleoperation, is handled via the Milesight UR75-500GL-G-P-W industrial cellular 5G router. 
The high transmission rate of the 5G standard is especially important for teleoperation to ensure low latency and high bitrates. 
The router is equipped with dual SIM cards for backup between multiple carrier networks; the router supports PoE and has an integrated GPS module.
As antennas, we selected the model LGMQM4-6-60-24-58 from Panorama Antennas, which supports 3G, 4G, 5G, GPS, and WiFi. 
We integrated three of them to operate all three SDR transceivers independently. 
Furthermore, \textit{EDGAR} has an additional coaxial connector on the roof to add further antennas for specific use cases.

\subsection{Actuators}
\label{sec:actuation}
The interfaces between the AV HPCs and the series actuators are realized via CAN. A vehicle gateway serves as an API between the AV commands and the series interface.

In addition, LEDs are placed around the roof of the vehicle. These serve as external Human-Machine-Interface (eHMI) and are controlled via a USB-interface. 

\begin{table*}
    \centering
    \caption{Specification of the components of the HiL simulator.}
    \label{tab:hil}
    \begin{tabular}{lllll}
    \toprule
    \textbf{Server} & \textbf{CPU} & \textbf{RAM} & \textbf{Storage} & \textbf{NVIDIA GPU}  \\
      \midrule
      GPU & 2x AMD EPYC 75F3, \SI{3.5}{\giga\hertz}  & 16x \SI{32}{\giga\byte} & 1x \SI{7.68}{\tera\byte} NVMe; 2x \SI{3.84}{\tera\byte} SAS  & 2x A40 \SI{48}{\giga\byte}  \\ 
      Visualization & Intel Xeon E-2286G, \SI{4.0}{\giga\hertz} & 4x \SI{8}{\giga\byte} & 1x \SI{1}{\tera\byte} NVMe & 2x A2000 \SI{12}{\giga\byte} \\
      \bottomrule
    \end{tabular}
\end{table*}

\begin{table*}
  \centering
  \setlength{\tabcolsep}{0.13cm}
  \begin{center}
    \caption{Specification of the components of the data center.}
    \label{tab:data_center_servers}
    \centering
    \begin{tabular}{llllll}
    \hline
    \textbf{Name} & \textbf{Qty} & \textbf{CPU} & \textbf{RAM} & \textbf{Storage} & \textbf{GPU}  \\
      \hline
      SiL & 1 & 2x AMD EPYC 75F3, \SI{2.95}{\giga\hertz} & 16x \SI{32}{\giga\byte} & 1x \SI{7.68}{\tera\byte} NVMe; 2x \SI{3.84}{\tera\byte} SAS  & -- \\ 
      x86 CI & 1 & 2x Intel Xeon Platinum 8362, \SI{2.8}{\giga\hertz} & 16x \SI{16}{\giga\byte} & 1x \SI{7.68}{\tera\byte} NVMe; 2x \SI{3.84}{\tera\byte} SAS & 2x A40 \SI{48}{\giga\byte}  \\
      ARM CI & 1 & 1 x Ampere Altra Max, M128-30, \SI{3.0}{\giga\hertz} & 2x \SI{64}{\giga\byte} & 2x \SI{7.68}{\tera\byte} NVMe; 2x \SI{0.96}{\tera\byte} SAS & 1x A16 \SI{64}{\giga\byte}  \\
      MAC CI & 1 & 1 x M2 Pro + 16-Core Neural Engine  & \SI{32}{\giga\byte} & 4x \SI{4}{\tera\byte} SSD & 19-Core GPU \\
      GPU & 1 & 2x AMD EPYC 9474F, \SI{3.6}{\giga\hertz}  &  16x \SI{128}{\giga\byte} &  3x \SI{7.68}{\tera\byte} NVMe &  2x A100 \SI{80}{\giga\byte}     \\
      HDD & 25 & 2x AMD EPYC 7252, \SI{3.1}{\giga\hertz} & 16x \SI{8}{\giga\byte} & 2x \SI{1.92}{\tera\byte} NVMe; 12x \SI{12}{\tera\byte}  SAS & -- \\
      SSD & 3 & 2x AMD EPYC 7313, \SI{3.0}{\giga\hertz}  & 12x \SI{16}{\giga\byte} & 16x \SI{7.68}{\tera\byte} NVMe; 2x \SI{1.92}{\tera\byte} NVMe & -- \\
      \hline
    \end{tabular}
  \end{center}
\end{table*}

\subsection{HiL simulator}
\label{sec:hil_sim}
\begin{figure*}
\centerline{\includegraphics[width=2.0\columnwidth]{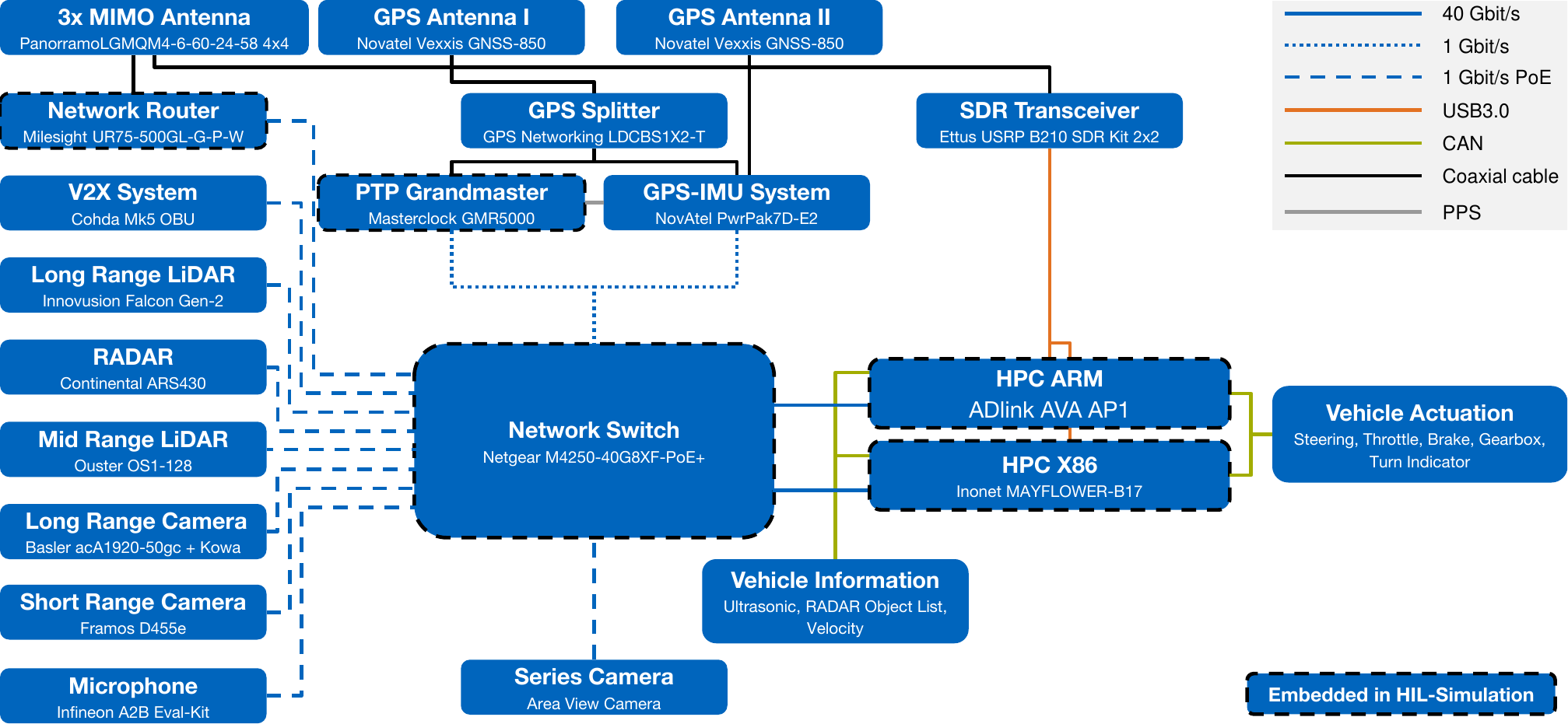}}
\caption{Network setup of the \textit{EDGAR} platform comprising sensors, network switch, HPCs, and vehicle interfaces. For the sake of simplicity, the gateway between the series sensors and the AV hardware and the gateway to the actuators are not shown in the figure. In addition, the components embedded into the HiL simulation are shown. \label{fig:network_setup}}
\end{figure*}

A custom HiL simulator is built to enable quick development cycles, software optimization, and interface testing. Using the same hardware as deployed in the real vehicle is crucial, as it ensures comparability of the results generated in the HiL simulator and simple transferability of software modules to the real vehicle.
The HiL setup comprises the same network router, switch, HPCs, and PTP GM (Fig. \ref{fig:network_setup}). Hence, the autonomous software on the HPCs can be evaluated, but it is also possible to evaluate the network setup and to run in teleoperated mode.
To ensure consistency between simulation and real-world tests, our digital twin is embedded into the HiL-simulation. Its specification is given in the repository, including vehicle parameters and sensor mounts.

The HiL simulator is primarily used for virtual validation tests before real-world tests.
In addition, with the virtual sensor setup placed into the simulation, it is possible to evaluate the autonomous software performance with the same constraints of occlusions and limited resolutions as the real-world vehicle. Also, the sensor settings can be analyzed, adjusted, and transferred to the real-vehicle setup.
Another critical use case of the HiL simulator is the generation of synthetic data:
These synthetic data sets generated in the simulation environment are crucial for developing perception algorithms.
The theoretically unlimited synthetic data allows the creation of a diverse data set that includes a wide range of traffic patterns and weather conditions and further enables the adaptation to different environments.
Additionally, hazardous scenarios can be simulated and included in the data set.
Since the ground-truth positions of all objects in the simulation are known, the labor-intensive and often manual labeling of real-world data is not needed for synthetic data.

Fig. \ref{fig:hil_setup} depicts our HiL simulator architecture. Table~\ref{tab:hil} outlines the specifications.
The GPU server runs the environment simulation, which includes virtual sensor models to generate synthetic sensor data and a vehicle dynamics model to simulate a sophisticated vehicle physics representation.
The AV software runs on the vehicle computers (x86, ARM), which have the same interfaces as in the real vehicle.
The ADLINK AVA Developer platform serves to develop code for arm-based CPUs efficiently. It is comparable to the AVA AP1 but has a lower number of cores (32) and clock rate (\SI{1.5}{\giga\hertz}). 
The visualization server is used as input to control the other servers and the AV software stack. Furthermore, it displays the graphical output of the environment simulation and AV software states.
All four compute platforms are connected via the network switch and communicate via ROS2.

\subsection{Data Center}
\label{sec:datacenter}
The \textit{EDGAR} data center consists of data storage and computing servers.
The data storage contains recorded sensor data from real-world test drives and simulations. 
The computing servers provide data access management, continuous integration (CI), SiL, training neural networks, and data analysis. 
We equipped the data center infrastructure with the servers listed in Table~\ref{tab:data_center_servers}.
The storage is separated into price-efficient hard drive disk (HDD) and latency-efficient solid-state drive (SSD) storage.
The HDD storage is used as the main storage component of the data, whereas the SSD storage is used for frequently-used data provided on demand for computation tasks.
Additionally, we have separate servers for CI tasks on different chips (x86, ARM), SiL, GPU-intensive tasks, and storage. 
The storage servers are integrated into an existing Ceph \cite{Weil2006} cluster, which provides redundancy for hardware failures. 
From the total amount of storage, approximately \SI{43}{\percent} is used by the Ceph cluster for redundancy.

Our data center aims to provide captured data for the development process in two stages:

First, we provide the raw recorded data in rosbags. These can be used for scenario replay, post-test scenario analysis, and scenario-oriented development.

\begin{figure}[b!]
\centering
\includegraphics[width=0.8\columnwidth]{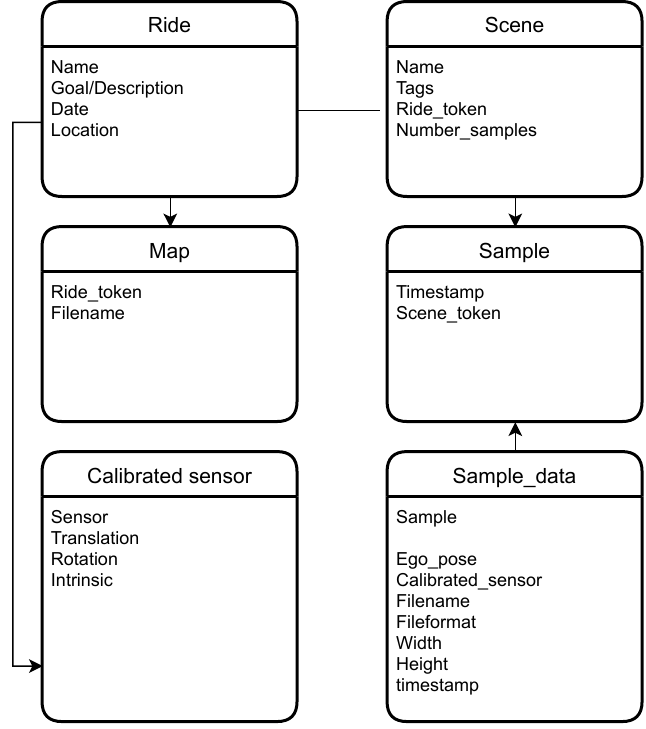}
\caption{Entity-Relationship Diagram. \label{fig:db-architecture}}
\end{figure}

Second, we extract and preprocess data from the available sensors and actuators embedded in the vehicle. This information collection includes images, point clouds, IMU and GPS readings, and CAN messages.
To present the data systematically and structured, we organize it into a relational database, similar to the NuScenes dataset \cite{nuscenes}. The architecture of this relational database is shown in Fig.~\ref{fig:db-architecture}.

\begin{figure}[t]
\includegraphics[width=1\columnwidth]{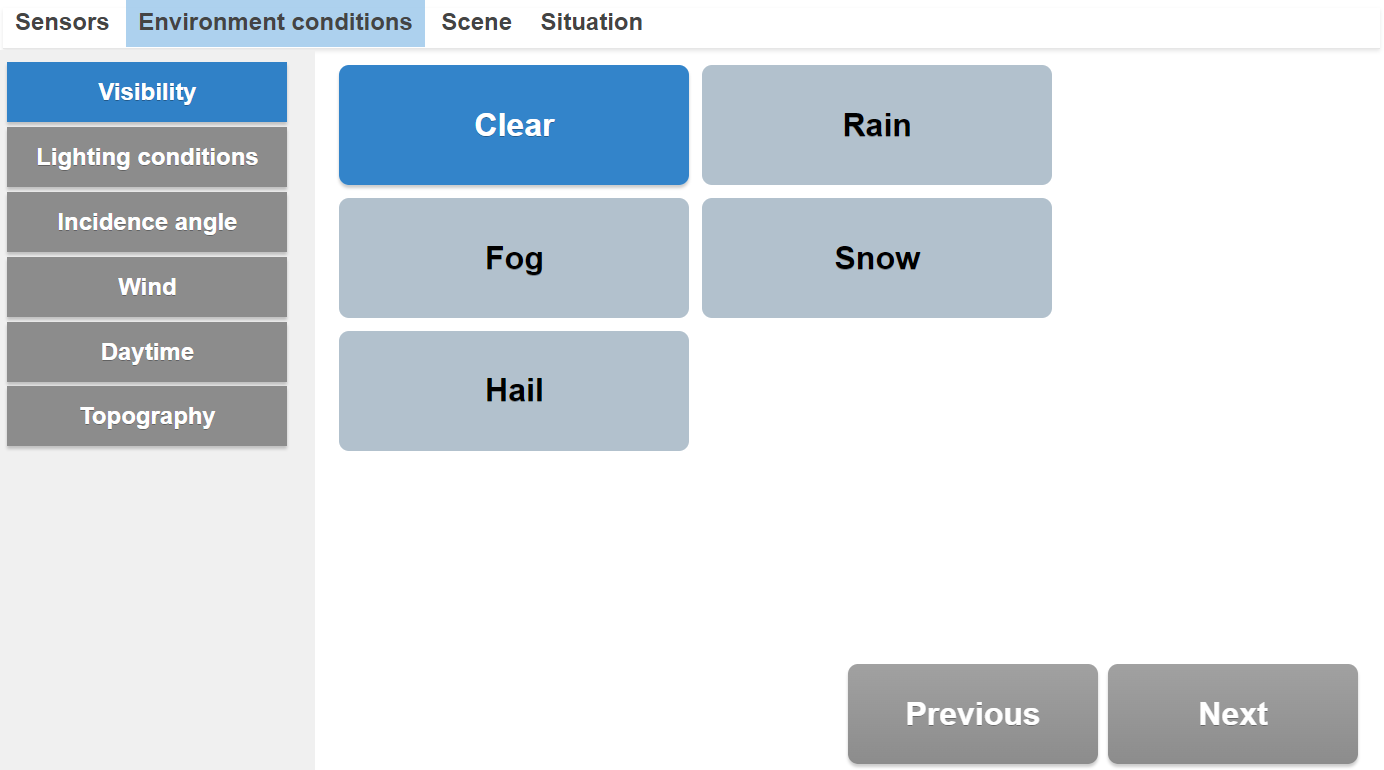}
\caption{Tagging interface with categories displayed in the top navigation bar, groups on the left side, and tags in the center of the screen\label{fig:tag-interface}.}
\end{figure}

Our top element is the \textit{ride}. It resembles a real-world ride while simultaneously recording raw data. Each \textit{ride} has a \textit{calibrated sensor} table associated with it, with the intrinsic and extrinsic calibration matrices of the sensors. Similarly, each ride gets assigned a \textit{map}. The ride is further divided into \textit{scenes} of specific duration, which are tagged to classify the situations of interest. Within each \textit{scene}, the time stamps for which all sensors have valid measurements are defined as a \textit{sample}. The \textit{sample data} are the measurements within the time window of the sample; each \textit{sample data} has an attribute indicating which sensor recorded the sample. Finally, for each \textit{sample}, an \textit{ego pose} for the vehicle is also measured.

The tagging interface employed to label each recorded scene on an abstract level is depicted in Fig. \ref{fig:tag-interface} and deployed within the vehicle. A hierarchical approach was chosen for a fast and efficient labeling process.
Each tag is assigned to a group and a category, while only relevant groups are 
displayed depending on the former selection. Furthermore, some tags, e.g., the sensor modalities or the vehicle speed, are automatically selected based on the information in the recorded rosbag.


\section{System Design}
\label{sec:system}
We now present the overall system design of the vehicle, which includes all aforementioned components (\ref{subsec:system_network}). Moreover, the driving modes to run the vehicle are introduced~(\ref{subsec:driving_modes}).

\subsection{Network}
\label{subsec:system_network}
The network, as shown in Fig. \ref{fig:network_setup}, comprises the environmental sensors, antennas, HPCs, and actuator interfaces.
The vehicle's mobile internet connection is established via the 5G antenna and the network router, which creates a VPN  to all components in the vehicle to keep them in an isolated network.
Both GPS antennas are input to the GPS-IMU system. However, one of the antennas is split and is also used as a reference clock for the PTP grandmaster. To synchronize the time of the GPS-IMU-system and the PTP grandmaster, an additional PPS signal is sent to the grandmaster. The GM sends its time as a master time to the network switch, which distributes it to all sensors and computers.

The core element of the system is the network switch, which receives all sensor data from the AV sensors and the series sensors and passes them through \SI{40}{\giga\bit\per\second} Ethernet to the two HPCs. 
The switch, as outlined, operates as a TC, which allows for the exchange of PTP messages and updating their residence time. 
In addition, ultrasonic, RADAR, and camera signals are received via CAN. 
The AV HPCs output a CAN signal to actuate the vehicle. 
There are interfaces for the steering wheel, throttle and brake pedal, gearbox, and turn indicator, among others.

For an In-Vehicle Network (IVN), we must also validate that network packets arrive at their destination within predefined time bounds.
Currently, no time bounds are ensured, but the system design allows for such guarantees, e.g., using the AVB/TSN features of the switch.
The traffic prioritization over other traffic ensures Quality of Service (QoS) for higher-priority traffic. 
This is especially important as combining all of the sensor data generates a large throughput, which, without proper management, can either overload the network or result in delays of high-priority/real-time traffic. 
The requirements on IVN were defined by the AVNU alliance~\cite{Automoti22:online} and categorized into Stream Reservation (SR) classes. 
The highest priority traffic belonging to the SR Class A requires a delay of less than \SI{2}{\milli\second} and a jitter of less than \SI{125}{\micro\second} over seven hops~ \cite{Automoti22:online}.

To determine what TSN configurations are required, we plan to collect the various data feeds from the sources.
Therefore, the selected sensors and other components support PTP, allowing for precise timestamping of the packets and enabling accurate traffic pattern analysis and data fusion on the application layer. 


\subsection{Driving Modes}
\label{subsec:driving_modes}
The vehicle can run in four different modes, which are:
\begin{enumerate}
    \item \textbf{Series Vehicle}: In this mode, the additional AV hardware is disconnected from the power supply and electronically separated from both series sensors and actuators.
    \item \textbf{Measurement driving}: All actuator interfaces are electronically separated, but the AV sensors and the HPCs are enabled. Thus, the mode can be used for data recording or running the software in ghost mode. 
    \item \textbf{Autonomous mode}: Lateral and longitudinal control is done by our software with limits of maximum speed, longitudinal acceleration and deceleration, lateral acceleration, and steering rate. This mode is used for test runs on public roads. The mode is implemented so that the safety driver can manually override the steering, brake, and acceleration commands.
    \item \textbf{High-dynamic mode}: This mode is used only in testing areas. This mode does not have limitations on longitudinal acceleration and deceleration, and lateral acceleration on the software side. Thus, speeds up to \SI{130}{\kilo\meter\per\hour} are possible, and the maximal steering rate can be used. 
\end{enumerate}


\section{Digital Twin}
\label{sec:digtwin}
Based on the autonomous vehicle setup presented in Section~\ref{sec:av_setup} and the system designed in Section~\ref{sec:system}, a digital twin is created to align the characteristics of the vehicle in the virtual and real environment. The digital twin comprises three aspects, namely the vehicle dynamics model (\ref{subsec:vehdyn}), replication of the sensor setup (\ref{subsec:sensorsetup}), and the replication of the network setup~(\ref{subsec:network}).

\subsection{Vehicle Dynamics}
\label{subsec:vehdyn}
An appropriate vehicle dynamics model is essential for the virtual development and validation of motion planning and control algorithms. Various models, such as single-track, double-track, multi-body models, and finite element simulations, exist to capture vehicle dynamics  \cite{Guiggiani2014}. However, selecting the right model involves balancing complexity and efficiency.
We adopt a dynamic nonlinear single-track model to account for essential dynamic effects, considering the combined slip of lateral and longitudinal tire forces, rolling resistance, and aerodynamic effects. To simulate lateral tire forces accurately, we use the Pacejka Magic Formula  \cite{pacejka1997magic}.

\subsection{Environment Sensors}

\label{subsec:sensorsetup}
For synthetic data generation, perception algorithm development, and validation tests, the exact replication of the sensor setup described in Section~\ref{subsec:sensors} is another important aspect of the digital twin.
The position and orientation of each sensor are measured in the reference system of the middle of the rear axle. The sensors' specifications (range, FoV, resolution) are given by the manufacturers. All parameters can be found in the repository. Based on these data, a 3D model of the VW Multivan is equipped with the sensors. The respective sensor models for each sensor modality are taken from given open-source solutions.

Sensor models can be separated into three categories: high-, medium-, and low-fidelity models. Low- and medium-fidelity models primarily rely on ground-truth object lists to generate sensor data or simulate sensor behavior. High-fidelity sensor models aim to simulate the underlying physical processes and their interaction with an available 3D environment and allow for higher data quality at the price of higher computational resource demand~ \cite{Schlager2020}.

With open-source simulation environments for AVs based on established game engines like UnrealEngine~4.26~\cite{Dosovitskiy2017} or Unity~\cite{AWSIM2023}, the included camera models represent the state of the art for generating camera data. High-fidelity LiDAR and RADAR models rely on ray casting to simulate electromagnetic wave propagation. The available RobotechGPULiDAR allows the simulation of solid state and mechanical LiDARs via customizable LiDAR patterns~ \cite{Robotec2023}. High-fidelity RADAR simulations currently only exist as stand-alone developments~ \cite{Holder2021, Thieling2021}. However, it is possible to implement a high-fidelity RADAR in open-source simulation environments. Microphone sensor models are currently not part of any open-source simulation framework for autonomous driving. Based on these given open-source implementations, further implementations of sensor models are intended to iteratively improve the digital replication of our sensor setup.

\subsection{Network}
\label{subsec:network}
Another aspect is to validate and improve the underlying network that supports communication between the system components.
The system must be robust and ensure deterministic data delivery under strict timing constraints.
As mentioned in Section~\ref{subsec:system_network}, the system is designed with a cyber-physical digital twin containing the same network components to understand the impact of the various data streams from all sensor components in the network.
To offer more versatility, the twin not only contains the sensors placed inside of the vehicle but also enables the data replay or modification of the data streams. 
Such an approach allows the simulation of additional scenarios that might not be present during the actual operation of the vehicle but might occur in various edge cases.
The underlying network should be robust enough to handle sudden changes.

\begin{figure}
\includegraphics[width=1.0\columnwidth]{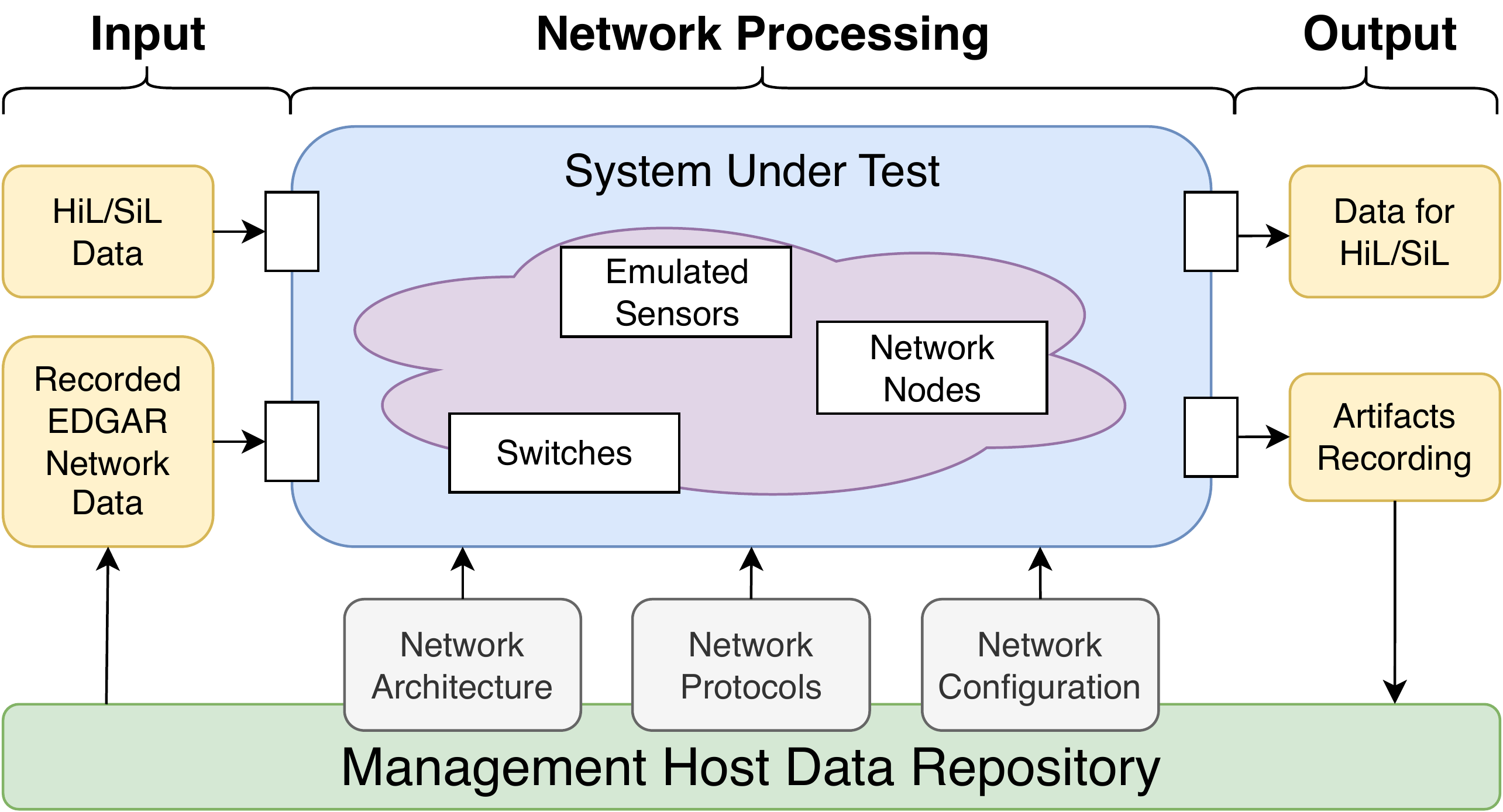}
\caption{EnGINE~ \cite{engine-jnsm} components for cyber-physical twin of \textit{EDGAR}\label{fig:engine-components}.}
\end{figure}

Our concept is to use the EnGINE framework~\cite{engine-jnsm, engine-method} combined with artifacts obtained from real-world \textit{EDGAR} testing.
The EnGINE framework is built using commodity off-the-shelf (COTS) hardware combined with open-source software solutions and enables verification of various network architectures and designs.
It supports the generation of synthetic traffic patterns and replay of collected packet traces in an experimental setup, built as shown in Fig.~\ref{fig:engine-components}.
EnGINE also enables AVB traffic shaping and PTP time synchronization, further supporting other IEEE 802.1Q Time-Sensitive Networking standards~ \cite{ieee8021q} relevant for AV, for example, IEEE 802.1Qav and Qbv standards.
Beyond its capability of serving as a HiL system representing a form of a cyber-physical twin, the framework is extended using simulation~\cite{engine-sim} based on the OMNeT++ discrete-event simulator.
In this way, EnGINE can also serve as a SiL tool, enabling the simultaneous execution of hardware-based and simulated experiments using a single configuration.

As a first step, using EnGINE we can build an exact representation of the network used within \textit{EDGAR} shown in Fig. \ref{fig:network_setup} centered around the Netgear M4250 network switch.
The real-world sensors will be emulated using collected artifacts and COTS hardware devices.
Such an approach improves the flexibility of the experimental environment while maintaining its realism and allows us to verify different protocols and the network configuration of \textit{EDGAR}.

\begin{figure*}
\centering
\includegraphics[clip, trim=0.6cm 2.1cm 0.8cm 2.2cm,width=2.0\columnwidth]{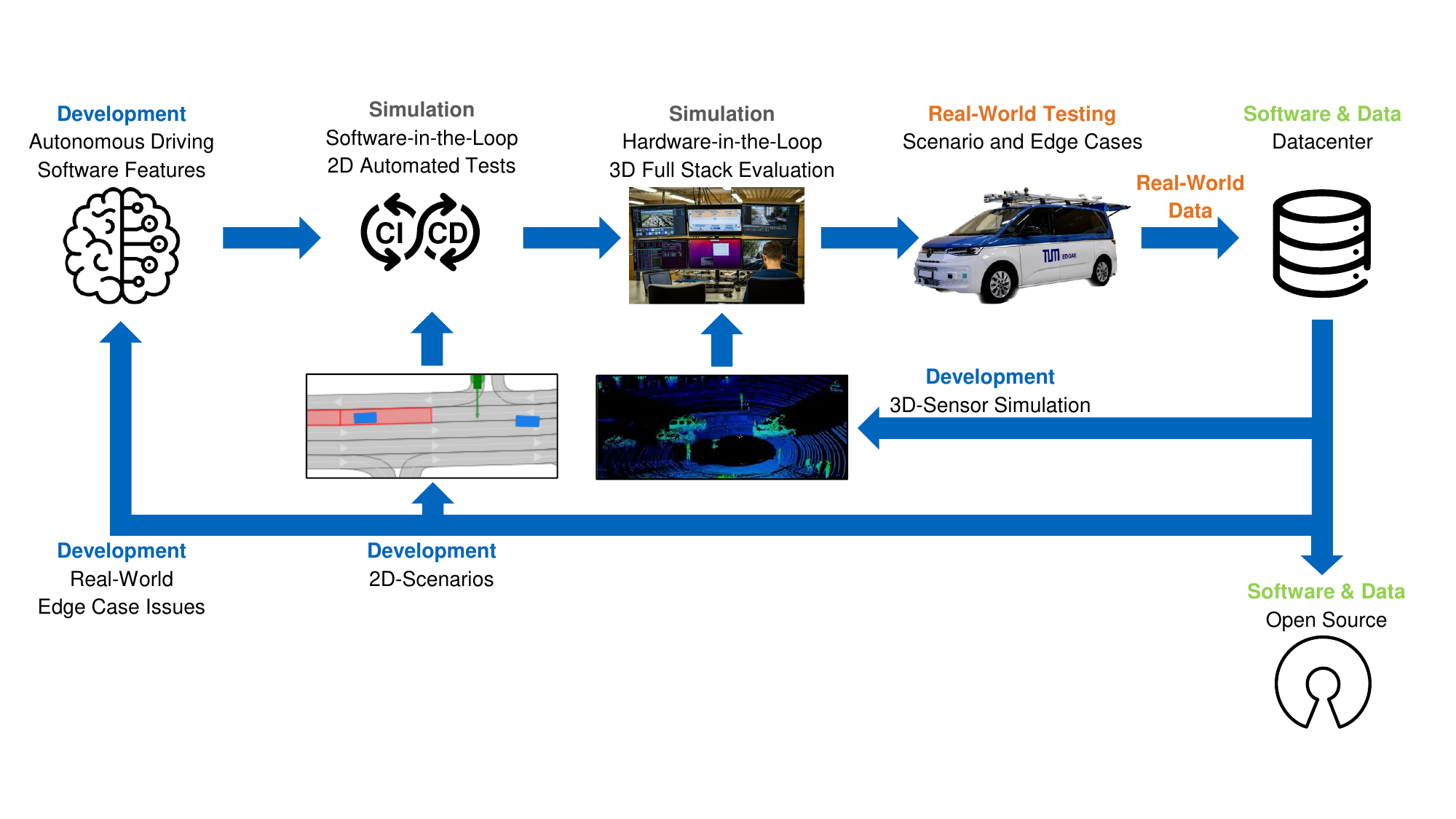} 
\caption{Development Workflow of \textit{EDGAR}.}
\label{fig:development_pipeline}
\end{figure*}

With its flexibility, EnGINE can serve as a platform to verify and improve \textit{EDGAR}'s network and its configuration.
The framework will allow us to focus on fulfilling the QoS requirements of various data streams by employing adequate TSN traffic shaping and policing mechanisms beyond PTP time synchronization.
With an understanding of the expected traffic patterns, we can use EnGINE to define and test those traffic shapers' appropriate selection and configuration.
For example, highly time-critical information would instead require the use of scheduling provided by the IEEE 802.1Qbv standard, while high-bitrate streams would benefit from the traffic shaping defined in the IEEE 802.1Qav standard.
EnGINE will allow us to define these correlations and ensure that \textit{EDGAR}'s IVN can support all QoS requirements of the interconnected devices.

In the future, EnGINE will also enable us to come up with novel network architectures, and designs can then be developed using a combined HiL and SiL approaches.
Using the framework's simulation extension, network topologies and device placements beyond what is currently available and deployed in \textit{EDGAR} can be initially evaluated in a SiL setup.
Such an approach can also enable further system optimization and shift focus toward the reliability and resilience of the IVN.
These can later be properly validated on the physical EnGINE setup before any changes to the AV architecture of \textit{EDGAR} are considered.


\section{Development Workflow}
\label{sec:dev_workflow}
The hardware setup presented in the previous sections needs to be applied in a suitable development workflow, which we present subsequently. A schematic overview of this workflow is shown in Fig. \ref{fig:development_pipeline}.

The feature development of our research covers aspects of every part of autonomous software.
The overall software architecture, into which the features are integrated, is \textit{Autoware Universe}~\cite{Kato2018}. Using this architecture, the developed code can be directly reused with other research institutions, and the open-source community's power significantly increases our development pace.

The developed features and the composed overall software are first evaluated in unit tests and 2D SiL simulations. These tests are part of the CI/CD toolchain, which is executed when the code is committed. In addition, tests are also executed in automated cloud-based scenario replays. Thereby, it is ensured that the software can be built and launched properly. Besides, the selection of standardized scenarios for the SiL-simulations allows us to track the progress of the overall software performance. Recorded sensor data are used to evaluate the performance of the perception modules, e.\,g. our concept DeepSTEP~\cite{Huch2023}. The motion planning benchmark framework CommonRoad is used as a scenario source to evaluate prediction and planning modules~\cite{Althoff2017}. It is chosen due to the diversity of more than 16,000 synthetic and real scenarios, which allows an objective evaluation of the implemented functions.
By means of the integration of CommonRoad, it is additionally possible to directly integrate the community planners used in the CommonRoad challenges~\cite{CRChallenges} and to use tools such as the integration in SUMO~\cite{klischat2019}, which encourages the modularity of the development environment.

The closed-loop full software stack simulation is the last step in the virtual test workflow. The software is deployed to the target computing platform and runs in a 3D environment. That is, all parts of the software are included in the tests. 

After the software passes the SiL- and HiL-test, the software is tested in real-world scenarios. To get as many insights as possible, our approach is to test the software in edge cases, i.e., at the limit of its capabilities. This contradicts a high distance record without disengagement, which is a common measure of the performance of AV software in the state of the art. However, the efficiency of our test procedure, in terms of new insights about the performance of our software per driven kilometer, is very high with this edge-case-driven approach.

The next step after the tests are conducted is data management. The selection of data that should be uploaded focuses on abnormal events. These events comprise scenarios that the software cannot solve, which are not covered by the simulation environment or are underrepresented in the data set for further development. From these abnormal events, 2D scenarios are extracted in case there are complex interactive scenarios that challenge the decision-making and motion planning algorithms, or in case of challenges to perceiving the environment, the data are passed to the 3D HiL simulation.
The 2D scenario extraction is implemented by a trajectory recording tool, which is directly compatible with the CommonRoad scenario library.

\begin{figure*}
\includegraphics[width=2.0\columnwidth]{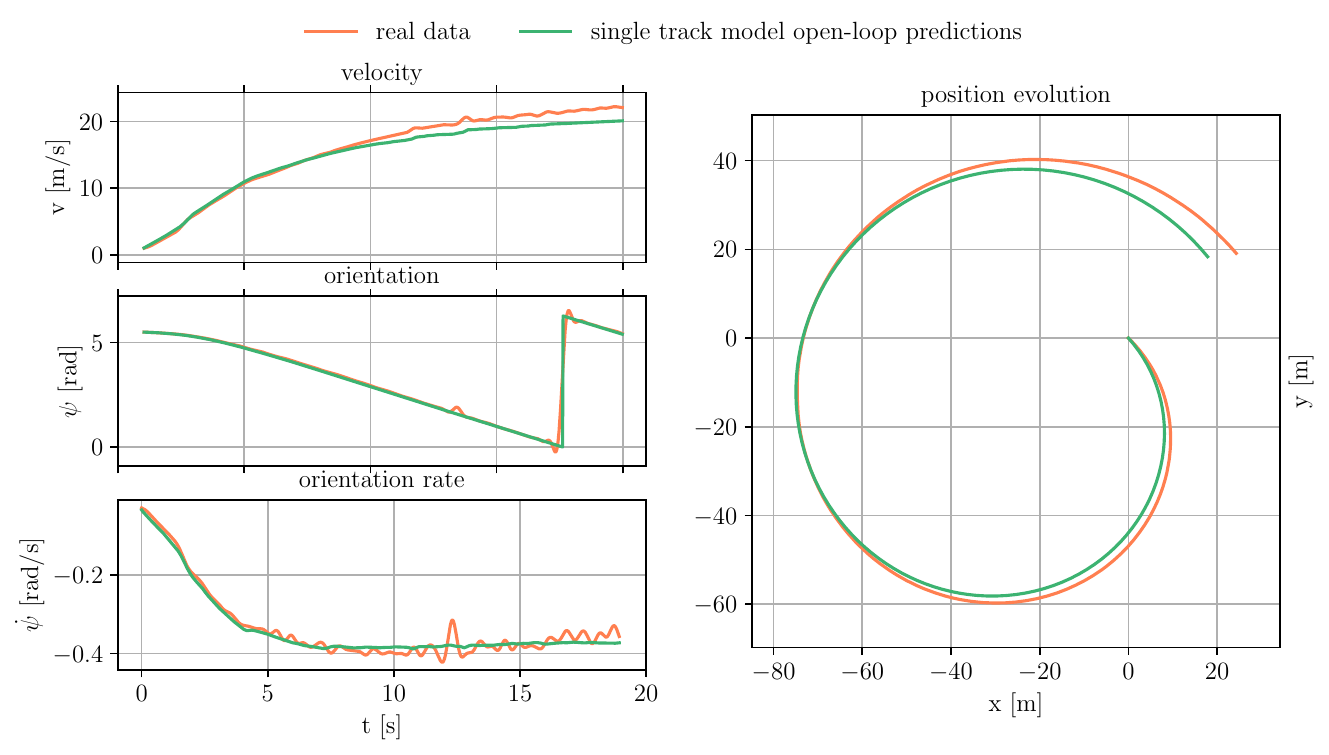}
\caption{Open-loop prediction assessment of the nonlinear single track model, parameterized as per Tab.\ref{tab:stm_parameters} and \ref{tab:tire_model}, against real measured data during a \SI{20}{\second} drive, equivalent to 1000 prediction steps. This evaluation is conducted during a continuous speed increase maneuver \cite{international2021passenger} with a constant steering wheel angle of \SI{90}{\degree} turning to the right \label{fig:open-loop-prediction} and with the starting position $(x,y)=(\SI{0}{\meter},\SI{0}{\meter})$.}
\end{figure*}

\section{Results} \label{sec:results}
In this section, the proposed vehicle concept (\ref{sec:av_setup}), which is duplicated into the Digital Twin (\ref{sec:digtwin}), is now validated in the real-world vehicle. A comparison is drawn regarding vehicle dynamics and sensor concept to the theoretical concept and to the virtual duplication to validate both effects. 
At the time of writing, the Digital Twin of the network is still in the early development stages and has not yet been validated but will be added to the open-source repository.

\subsection{Vehicle Dynamics}
We define the vehicle state vector, $\boldsymbol{x} \in \mathbb{R}^{n_x}$, and the control input vector,  $\boldsymbol{u} \in \mathbb{R}^{n_u}$, as follows:
\begin{equation}
\begin{aligned}
\boldsymbol{x} &= [x_{\text{pos}},\space y_{\text{pos}},\space \psi,\space v_{\text{lon}},\space v_{\text{lat}},\space \dot{\psi}]^T \\
\boldsymbol{u} &= [\space \delta_f,\space a ]^T
\end{aligned}
\label{eq:stm_states}
\end{equation}
Here, $x_{\text{pos}}$ and $y_{\text{pos}}$ denote the $x$- and $y$-coordinates of the ego vehicle, $\psi$ the yaw angle, $v_{\text{lon}}$ and $v_{\text{lat}}$ the velocities in the longitudinal and lateral directions, $\dot{\psi}$ the yaw rate, $\delta_f$ the steering angle at the front wheel, and $a$ the acceleration.

Validating the chosen model with real-world data and identifying parameter values are crucial for ensuring accuracy and reliability. We measure some parameters, including the position of the center of gravity and vehicle mass. The vehicle configuration employed in this study comprises five mounted seats and accommodates a driver and no passengers.

To identify further parameters, we conduct steady-state circular driving behavior tests compliant with ISO 4138  \cite{international2021passenger}. Our focus is on the constant steering-wheel angle approach, and we employ two variations: discrete- and continuous speed increase tests. We collect motion and steering data from the GPS-IMU and VW Series motion sensors.
To cover both low- and high-velocity ranges, we conduct tests at velocities from \SI{5}{\kilo\meter\per\hour} up to \SI{130}{\kilo\meter\per\hour}, with steering wheel angles ranging from \SI{45}{\degree} to \SI{540}{\degree} in both turning directions. Under normal circumstances, i.\,e., clear weather, negligible wind speed, and an outside temperature of \SI{23}{\degreeCelsius}, we use Bridgestone 235/50R18 101H summer tires.

\begin{table}[!b]
\caption{Identified single track parameters.}
\centering
\begin{tabular}{llll}
\toprule
& \textbf{Value} & \textbf{Unit} & \textbf{Description} \\
\midrule
$l      $    & 3.128 & \unit{\meter} & Wheelbase \\
$l_{\mathrm{f}}    $     & 1.484 & \unit{\meter}    & Front axle to center of gravity\\
$l_{\mathrm{r}}    $     & 1.644 & \unit{\meter}   & Rear axle to center of gravity\\
$m     $     & 2520  & \unit{\kilogram}   & Vehicle mass \\
$I_{\mathrm{z}}  $      & 13600 & \unit{\kilogram\meter\squared} & Moment of inertia in yaw \\
$\rho  $       & 1.225 & \unit{\kilogram\per\meter\cubed}    & Air density \\
$A  $        & 2.9   & \unit{\meter\squared}    & Cross-sectional frontal area \\
$c_{\mathrm{d}}$         & 0.35  &      & Drag coefficient \\
\bottomrule
\end{tabular}
\label{tab:stm_parameters}
\end{table}

Our goal is to estimate the single track model parameters that minimize the deviation between a series of true measured states $\boldsymbol{X_\text{true}} \in  \mathbb{R}^{n_s} \times \mathbb{R} ^{n_x}$ and the open-loop predicted 
states $\boldsymbol{X_\text{pred}} \in  \mathbb{R}^{n_s} \times \mathbb{R} ^{n_x}$, starting from the same initial state $\boldsymbol{x_\text{true}}^{(0)} = \boldsymbol{x_\text{pred}}^{(0)}$ and applying the same series of control vector inputs $\boldsymbol{U} \in  \mathbb{R}^{n_s} \times \mathbb{R} ^{n_u}$ for $n_s$ prediction steps. Here, $\boldsymbol{X_\text{true}} = [\boldsymbol{x_\text{true}}^{(1)}, \ldots, \boldsymbol{x_\text{true}}^{(n_s)}]$, $\boldsymbol{X_\text{pred}} = [\boldsymbol{x_\text{pred}}^{(1)}, \ldots, \boldsymbol{x_\text{pred}}^{(n_s)}]$, and $\boldsymbol{U} = [\boldsymbol{u}^{(0)}, \ldots, \boldsymbol{u}^{(n_s-1)}]$, where $\boldsymbol{x_\text{true}}^{(i)}$,  $\boldsymbol{x_\text{pred}}^{(i)}$, and $\boldsymbol{u}^{(0)}, \forall i \in \{0,\dots,n_s\}$ are defined as in Equation~\ref{eq:stm_states}. Our model operates on a discretization time of $T_s = \SI{0.02}{\second}$, and it is essential to highlight that the prediction model remains uninformed about the real states, i.\,e. the model does not receive updates during the simulation.

We refine our parameter estimates through an iterative process of manual tuning. The resulting single-track and tire model parameters are listed in Table \ref{tab:stm_parameters} and \ref{tab:tire_model}, respectively, where $F_{z,\{f/r\}}$ represents the vertical static tire load at the front and rear axles.

\begin{table}[!t]
\caption{Pacejka tire model parameters.}
\centering
\begin{tabular}{l l l l}
\toprule
\textbf{Parameter} & \textbf{{Front}} & \textbf{{Rear}} & \textbf{Description} \\
\midrule
$B$ & 10 & 10 & Stiffness factor \\
$C$ & 1.3 & 1.6 & Shape factor \\
$D$ & 1.2$\cdot F_{z,f}$ & 2.1$\cdot F_{z,r}$& Peak value \\
$E$ & 0.97 & 0.97 & Curvature factor \\
\bottomrule
\end{tabular}
\label{tab:tire_model}
\end{table}

In Figure \ref{fig:open-loop-prediction}, we present the open-loop prediction assessment spanning $n_s = 1000$ prediction steps, corresponding to a \SI{20}{\second} ride. During this evaluation, the steering-wheel angle remains fixed at \SI{90}{\degree}, while the vehicle speed undergoes a steady increase, as detailed in \cite{international2021passenger}. Table~\ref{tab:stm_pred_eval} complements this analysis, featuring the model's prediction accuracy evaluated through key metrics: Root Mean Square Error (RMSE), Mean Absolute Percentage Error (MAPE), and the coefficient of determination, R².

It is important to emphasize that the accuracy of parameter estimation relies heavily on both the selected model and the quality of the measured data. Additionally, our approach did not employ any state estimation and sensor fusion techniques to enhance state estimations.

\subsection{Sensor Concept}
The sensor concept evaluation comprises two focus points. At first, the sensor concept is qualitatively analyzed regarding the sensor coverage in a real-world scenario. Second, the influences of calibration and synchronization on the sensor performance are discussed. While the current state of the art focuses on developing sophisticated sensor models of a single modality, we analyze the Sim2Real gap on the system level.

\subsubsection{Sensor Coverage}
The sensor coverage is qualitatively evaluated in a real-world scenario of the target ODD of dense urban traffic. The scenario gives insights into the actual sensor coverage for the intended application.

\begin{figure}[!b]
\includegraphics[width=1.0\columnwidth]{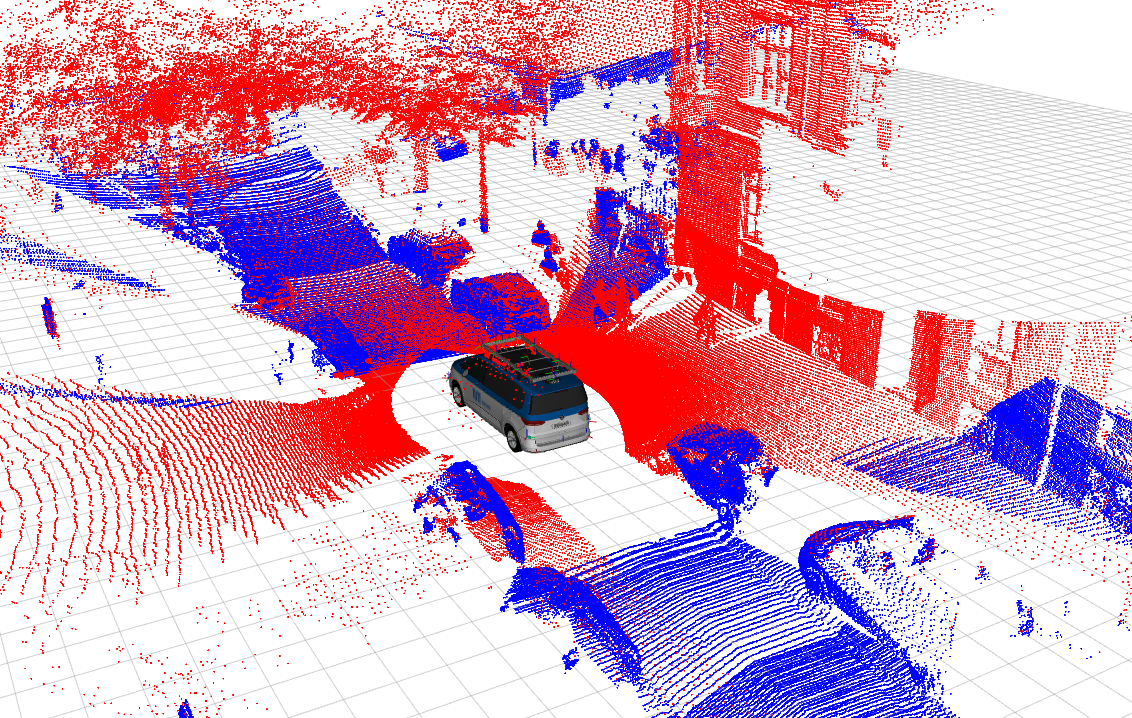}
\caption{Sensor coverage of the real-world vehicle in a populated environment (Ouster: Red, Innovusion: Blue).\label{fig:LiDAR_real}}
\end{figure}

Figure~\ref{fig:LiDAR_real} presents a visualization of the LiDAR coverage, with Ouster sensor detections marked in red and Innovusion detections in blue. The vehicle is located within a densely populated and narrow one-way street with a number of traffic participants, including pedestrians, cyclists, and parked vehicles on either side from the ego perspective, which presents a particularly challenging scenario for autonomous vehicles.

Our sensor setup aligns with the original conceptual design, wherein the red LiDAR points registered by the Ouster sensors form the oval near-field horizon typical of 360° LiDAR sensors. Meanwhile, the blue LiDAR points detected by the Innovusion sensors show the coverage both in the front and rear directions. The concatenation of data from all four sensors and considering their respective positions and orientations results in the emergence of near-field blind spots, as previously illustrated in Figure \ref{fig:lidar_unity}.

\begin{table}[t]
\centering
\caption{Assessing the predictive accuracy of the nonlinear single-track model through open-loop simulations, compared with real measured data obtained from the experimental drive illustrated in Fig. \ref{fig:open-loop-prediction}.}
\label{tab:stm_pred_eval}
\begin{tabular}{@{}llll@{}}
\toprule
\textbf{} & \textbf{RMSE} & \textbf{MAPE} & \textbf{R²} \\ \midrule
velocity $v$ & \SI{1.15}{\meter\per\second} & 5.71\% & 0.96 \\
yaw $\psi$& \SI{0.24}{\radian} & 5.80\% & 0.98 \\
yaw rate $\dot\psi$ & \SI{0.015}{\radian\per\second} & 4.71\% & 0.97 \\ \bottomrule
\end{tabular}
\end{table}


An additional point of consideration relates to the choice of \textit{not} utilizing a rotating LiDAR mounted on the roof that solely covers a 360° FoV. This decision results in a situation where the two Innovusion LiDAR sensors do not share a direct overlap in their fields of view.
This factor assumes considerable importance when assessing the extrinsic calibration between the LiDAR sensors. Given that the Innovusion and Ouster sensors exhibit partial overlap in their fields of view, achieving an optimal extrinsic joint calibration across all LiDAR units becomes essential.

\subsubsection{Sensor Calibration and Synchronization}
\begin{figure}[!b]
\includegraphics[width=1.0\columnwidth]{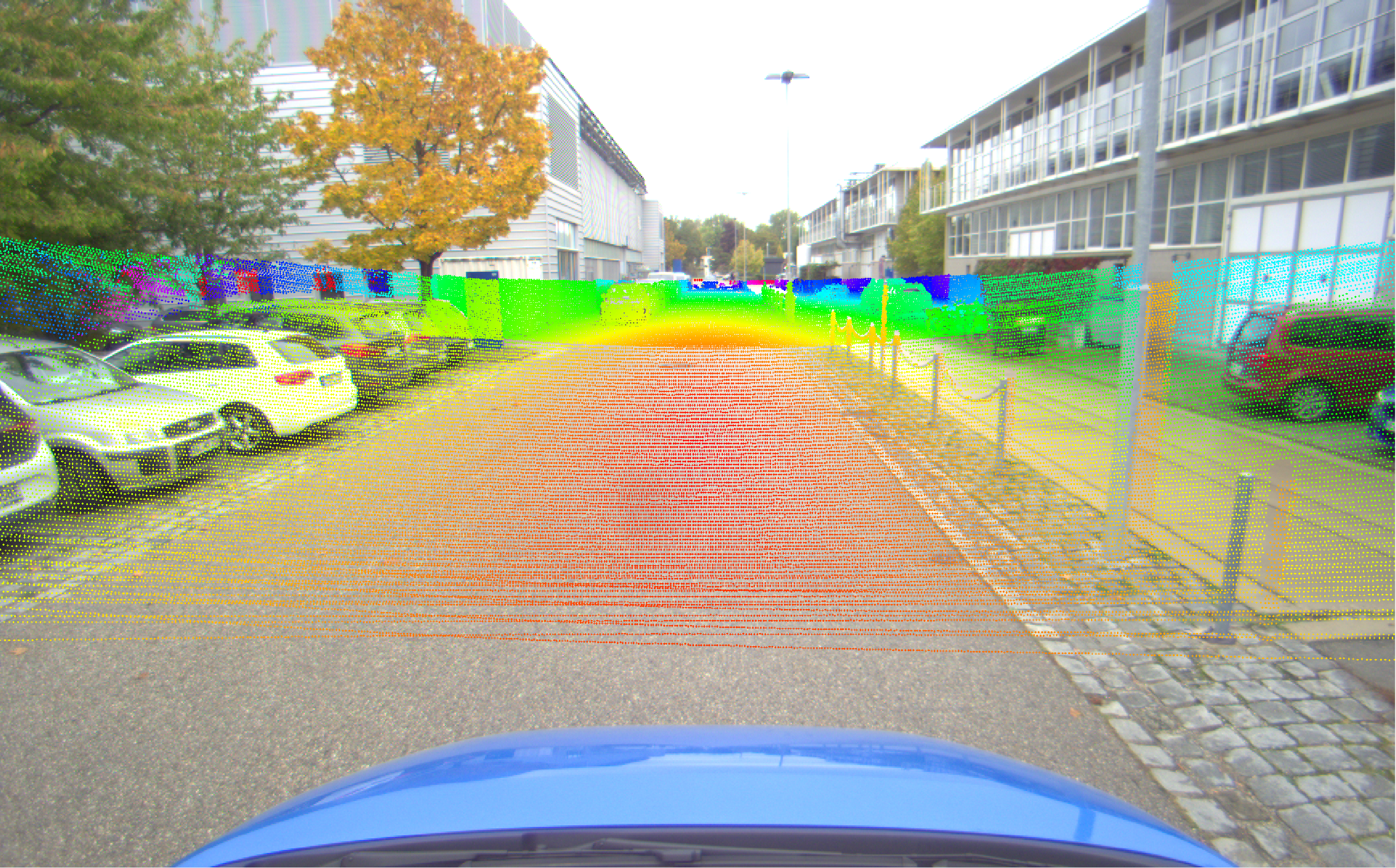}
\caption{Camera-LiDAR Projection.}
\label{fig:projection}
\end{figure}
In order to integrate data from both camera and LiDAR sensors within a real-world setup, a crucial step is the calibration and synchronization process. We employed Koide's calibration tool~\cite{koide2023}, which requires a static recording of the sensors to generate the transformation matrix connecting the camera and LiDAR. To assess the quality of this procedure, we project the LiDAR's point cloud onto the camera's field of view, as illustrated in Figure~\ref{fig:projection}.

\begin{figure}[!t]
\includegraphics[width=1.0\columnwidth]{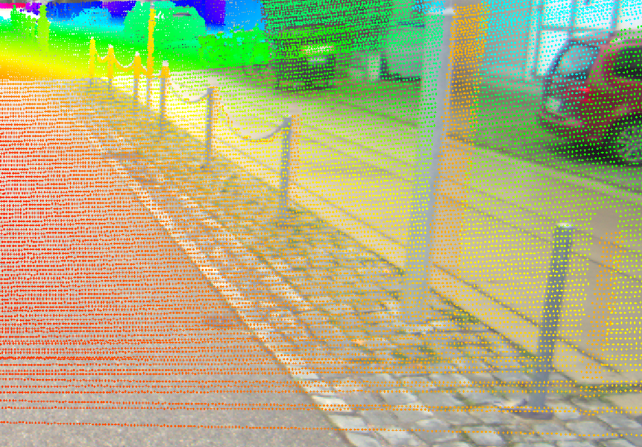}
\caption{Projection snippet at 0 \si[text-series-to-math]{\kilo\metre\per\hour}. Projection error due to camera distortion is \SI{2.9}{\percent} on the pole at the right border of the picture.}
\label{fig:projection_0}
\end{figure}

We designed three scenarios to assess the impact of synchronization and calibration. First, to exclude dynamic effects, we present a recording at \SI{0}{\kilo\metre\per\hour} in Figure \ref{fig:projection_0}. This scenario reveals that the calibration performs optimally in the central region of the field of view but exhibits distortion at the edges. Thus, the intrinsic calibration matrix of the camera is not optimally set.
The relative projection error of the camera is evaluated on the characteristic point of the most right pole in Figure~\ref{fig:projection_0}. It is determined by the distance between the LiDAR point cloud projection and the camera image at the characteristic point of the pole. The distance is set in relation to the camera's horizontal resolution to obtain the relative projection error inside the camera's horizontal FoV. In the static case (Figure~\ref{fig:projection_0}), the relative projection error is \SI{2.9}{\percent}.

\begin{figure}[!t]
\includegraphics[width=1.0\columnwidth]{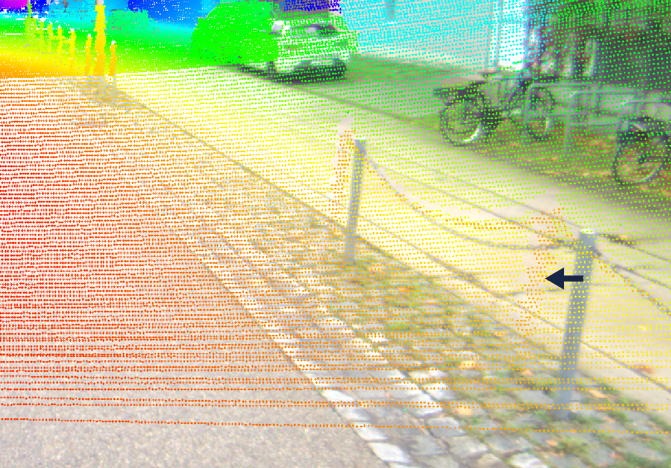}
\caption{Projection snippet at 30 \si[text-series-to-math]{\kilo\metre\per\hour}. Projection error due to camera distortion and unsynchronised sensors is \SI{5.2}{\percent} on the pole at the right border of the picture.}
\label{fig:projection_30}
\end{figure}

\begin{figure}[!t]
\includegraphics[width=1.0\columnwidth]{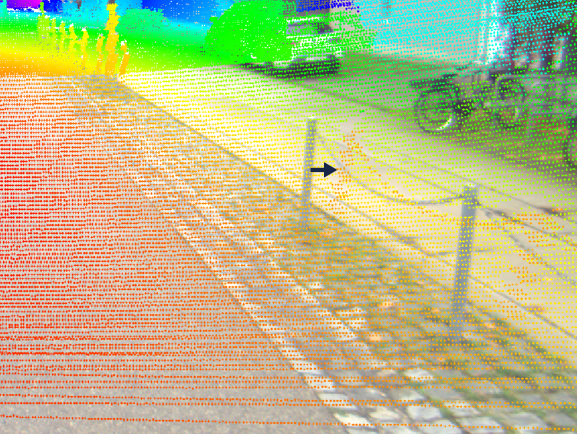}
\caption{Projection snippet at 50 \si[text-series-to-math]{\kilo\metre\per\hour}. Projection error due to camera distortion and unsynchronised sensors is \SI{6.8}{\percent} on the pole at the right border of the picture.}
\label{fig:projection_50}
\end{figure}

Next, to evaluate the synchronization behavior, we provide a recording at \SI{30}{\kilo\metre\per\hour} in Figure \ref{fig:projection_30}. In this case, the misalignment at the image edges between the camera and LiDAR is more pronounced compared to the static case.
The value for the projection error is \SI{5.2}{\percent} at the edge of the camera's FoV. 
Besides the distortion error, analyzed in the static case, additional disparities in triggering between the sensors exist. The sensors are not synchronously triggered, resulting in timestamps separated by up to \SI{25}{\milli\second}, corresponding to a \SI{20}{\centi\metre} travel distance between camera and LiDAR frames. Given the analysis from Figure~\ref{fig:projection_0}, the isolated error due to unsynchronized sensors is around \SI{2.3}{\percent} at this speed.
This issue could be resolved by simultaneously triggering both sensors. However, the chosen sensors do not offer this feature. An increase in the sampling rate could at least reduce the effect of the temporal disparity.

Finally, we recorded a scenario at \SI{50}{\kilo\metre\per\hour} to examine the sensitivity of the synchronization error against the ego speed. The results are depicted in Figure \ref{fig:projection_50}.
The misalignment further increases around the pole. Even in the image's central region, a slight misalignment between the point cloud and the image is noticeable. The projection error increases to \SI{6.8}{\percent} at the edge of the camera's FoV. Thus, the synchronization error increases linearly with the ego speed in this speed range.
Furthermore, a rolling shutter effect is observed around slender objects due to inherent triggering delays in the LiDAR points.

These effects serve to illustrate the domain gap that emerges when comparing a real vehicle implementation with its digital twin on the system level. As a result, we outline the factors contributing to this disparity: image distortion and unsynchronized sensor triggering. In addition, the rolling shutter of the LiDAR sensor influences the point cloud alignment, starting from the speed of \SI{50}{\kilo\meter\per\hour}.


\section{Conclusion}
A holistic platform for autonomous driving research is introduced. The core element is our research vehicle \textit{EDGAR} and its digital twin, a virtual duplication of the vehicle. The vehicle is equipped with a multi-modal state of the art sensor setup, HPC platforms with different chip technologies, and fully accessible actuator interfaces. Its digital twin comprises vehicle dynamic models and sensor and network duplicates for consistency between virtual and real-world testing. To the best of our knowledge, this is the first publicly available digital twin of an autonomous road vehicle. It ensures consistency between virtual and real-world tests, facilitates deployment, and reduces the integration effort of new software features. All of these aspects boost the development of AV software stacks. 
The real and virtual vehicles embedded in the presented development workflow with a multi-stage simulation and testing approach and a large-scale data center. The proposed workflow covers the full process from feature development up to full-stack real-world tests.
The results reveal the discrepancy between the digital twin and the real-world behavior. In terms of vehicle dynamics, the open-loop validation reveals a drift in speed and position estimation while the orientation is accurately modeled. The validation of the sensor concept reveals a trade-off in sensor coverage between range and occluded areas. The aspects of calibration and synchronization are examined and reveal a gap between simulation and reality on the system level caused by distortions, missing sensor triggering, and rolling shutter. Besides the development of sophisticated sensor models for single modalities, there is a need for research on sensor system level.

Future work will tackle three central aspects. At first, we will validate the efficacy of our development process around the digital twin. It shall be investigated if the validation framework is able to prove the functionality on an algorithm, module, and overall software level. Our goal is to continuously improve the virtual validation stages by comparing the real-world performance of the software with the evaluation at the different simulation stages. The incomplete simulation behavior of our digital twin shall be corrected based on real-world observations. Thus, our digital twin will be continuously improved.
Second, based on the introduced research platform, the development of new module software features, simulation models, evaluation frameworks, and software stack optimization is intended. In contrast to other works, our developed methods will always be validated within the whole software stack to analyze the dependencies and performance in a full stack.
Lastly, we aim to create a large-scale urban, multi-modal data set. With a focus on edge cases such as adverse weather conditions and abnormal behavior of traffic participants and using auto-labeling and anomaly detection tools, we want to achieve a diverse data set to foster future research and software development.

All parts of our developed software and all collected data will be published open-source to share the knowledge and insights gained with the research community to accelerate the progress in autonomous driving research.







\section*{CONTRIBUTIONS}
As the first author, Phillip Karle initiated the idea of this paper, created the overall structure, and essentially contributed to all sections of the paper. The other authors contributed to the sections on the autonomous vehicle setup, system design, digital twin, development workflow, and the overall research projects.
Johannes Betz contributed to essential parts of the paper and to the conception of the DFG proposal.
Matthias Althoff led the DFG proposal for financing the vehicle; he had the idea to develop the digital twin. In addition, he developed the concept of sharing all data in a common data center and leads the CommonRoad project and its integration into EDGAR.
Markus Lienkamp made an essential contribution to the conception of the DFG proposal. He supervised the setup of the vehicle and HiL and the conception of the development and validation workflow. He revised the paper critically for important intellectual content. He gave final approval of the version to be published and agreed with all aspects of the work. As a guarantor, he accepts responsibility for the overall integrity of the paper.

\section*{ACKNOWLEDGEMENTS}
The vehicle was partly sponsored by a DFG grant (approval according to Art. 91b GG with DFG-number INST 95/1653-1 FUGG). In addition, the project is supported by the Bavarian Research Foundation (BFS), by MCube - Munich Cluster for the Future of Mobility in Metropolitan Regions, the German Research Community (DFG), the Federal Ministry for Economics Affairs and Climate Action, and by the research project ATLAS L4.

We gratefully thank our partners, Arm and the Xilinx University Program, for the donation of hardware platforms for our research environment.

\bibliographystyle{IEEEtran}
\bibliography{main}





\end{document}